\PassOptionsToPackage{table}{xcolor}
\documentclass[acmtog,nonacm]{acmart}

\usepackage{geometry}
\usepackage{multirow}
\usepackage{colortbl}
\usepackage{pifont}
\usepackage{booktabs}
\usepackage{graphicx}
\usepackage[table]{xcolor}
\usepackage{subcaption}
\usepackage{amsmath}
\usepackage{algorithm}
\usepackage{algpseudocode}
\usepackage{float}

\definecolor{colorfirst}{HTML}{ffb2b3}  
\definecolor{colorsecond}{HTML}{ffd9b6} 
\definecolor{colorthird}{HTML}{ffffb9}  

\AtBeginDocument{%
  }

\setcopyright{acmlicensed}
\copyrightyear{2026}
\acmYear{2026}
\acmDOI{XXXXXXX.XXXXXXX}

\acmJournal{TOG}

\begin{document}

\title{TopoSurfel: Closing the Loop between Gaussian Surfels and Meshes for Surface Reconstruction}

\author{Chuanjin Fan}
\orcid{0009-0003-8414-318X}
\affiliation{%
  \institution{University of Science and Technology of China}
  \city{Hefei}
  \country{China}
}
\email{fancj@mail.ustc.edu.cn}

\author{Wenjie Chang}
\orcid{0009-0006-5603-6936}
\affiliation{%
  \institution{University of Science and Technology of China}
  \city{Hefei}
  \country{China}
}
\email{changwj@mail.ustc.edu.cn}

\author{Bohao Liao}
\orcid{0009-0007-5020-377X}
\affiliation{%
  \institution{University of Science and Technology of China}
  \city{Hefei}
  \country{China}
}
\email{liaobh@mail.ustc.edu.cn}

\author{Yujia Chen}
\orcid{0009-0005-1533-9263}
\affiliation{%
  \institution{University of Science and Technology of China}
  \city{Hefei}
  \country{China}
}
\email{yujia_chen@mail.ustc.edu.cn}

\author{Wenfei Yang}
\orcid{0000-0003-3599-7659}
\affiliation{%
  \institution{University of Science and Technology of China}
  \city{Hefei}
  \country{China}
}
\email{yangwf@ustc.edu.cn}

\author{Tianzhu Zhang}
\orcid{0000-0003-0764-6106}
\affiliation{%
  \institution{University of Science and Technology of China}
  \city{Hefei}
  \country{China}
}
\affiliation{%
  \institution{National Key Laboratory of Deep Space Exploration, Deep Space Exploration Laboratory}
  \city{Hefei}
  \country{China}
}
\email{tzzhang@ustc.edu.cn}

\renewcommand{\shortauthors}{Fan et al.}

\begin{abstract}
3D Gaussian Splatting has achieved remarkable success in novel view synthesis. However, extracting high-fidelity surfaces directly from 3DGS remains challenging due to its discrete and unstructured nature. Existing 3DGS-based reconstruction methods typically rely on multi-view geometric consistency or local constraints. Without an explicit structured geometric prior during optimization, these methods often struggle to resolve structural ambiguities, leading to artifacts and floaters, particularly in textureless or occluded regions. To address this limitation, we propose TopoSurfel, a novel framework that closes the loop between Gaussian surfels and continuous meshes. Unlike recent methods that incorporate mesh extraction into the differentiable pipeline by introducing auxiliary neural networks or extra per-Gaussian parameters, we dynamically extract a continuous proxy mesh via a non-trainable differentiable iso-surfacing process. Leveraging this differentiable connection, we introduce a mesh-guided surfel evolution strategy, including normal alignment and geometry-aware density control, to effectively suppress floaters and fill surface holes. Furthermore, to address the initialization challenges in large-scale environments, we propose a spatially aware hybrid re-initialization strategy that ensures robust reconstruction across complex scenes. Extensive experiments demonstrate that TopoSurfel achieves competitive geometric reconstruction accuracy while maintaining high-quality mesh-based novel view synthesis. The code for our method is available at \url{https://github.com/Fan-Treasure/TopoSurfel}.
\end{abstract}




\begin{teaserfigure}
  \Description{Overview of our surface reconstruction framework.}
  \centering
  \includegraphics[width=0.92\textwidth]{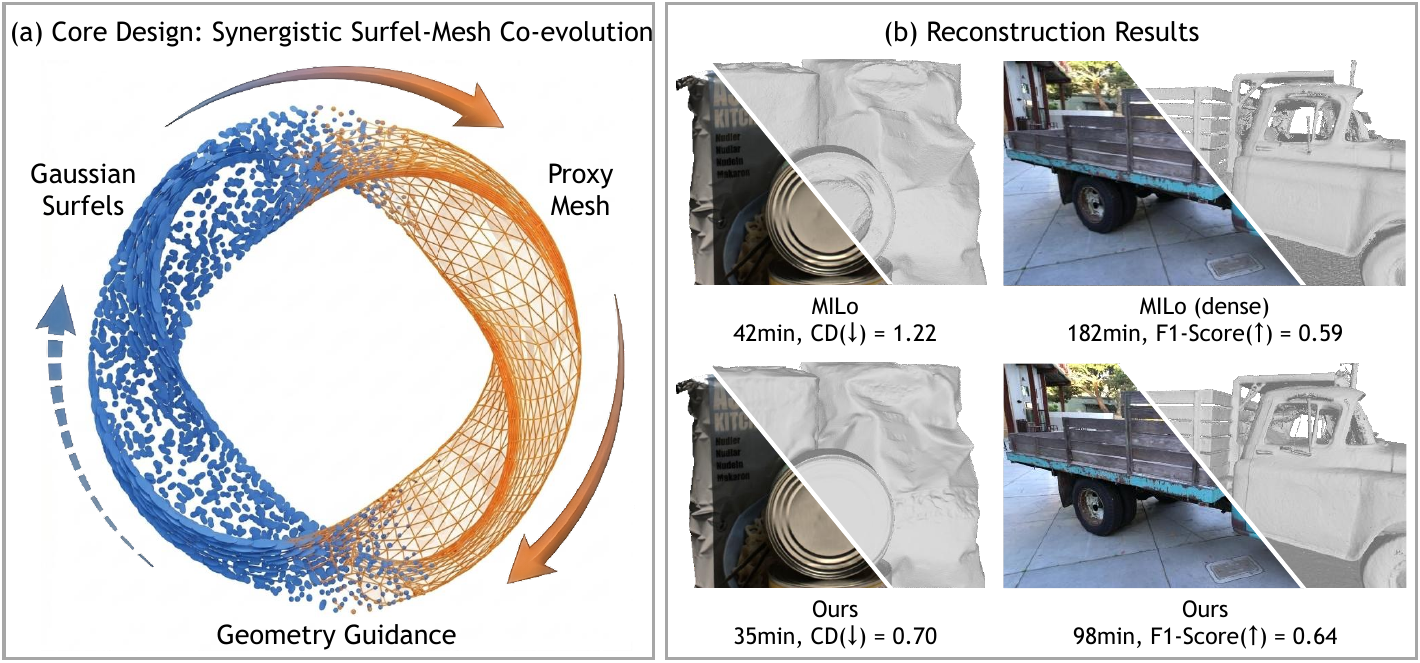}
  \caption{Overview of our surface reconstruction framework. (a) We propose a synergistic co-evolution scheme that bridges Gaussian surfels and differentiable meshes. By enforcing mutual geometry guidance, our method effectively captures both high-frequency details and coherent surface structures. (b) Reconstruction results on various scenes demonstrate that our method outperforms existing methods in both quality and efficiency.}
  \label{fig:teaser}
\end{teaserfigure}

\maketitle

\section{Introduction}
\label{sec:intro}

High-fidelity surface reconstruction lies at the core of 3D vision and graphics. It underpins applications in physics simulation, digital twins, material editing, and virtual reality. In recent years, 3D Gaussian Splatting \cite{kerbl2023gaussian} has achieved remarkable progress in novel view synthesis. However, 3DGS is a discrete and unstructured volumetric representation, which makes it difficult to reliably produce high-quality meshes.

To improve the geometric quality of Gaussian representations, a number of strong surface reconstruction methods have been proposed, including SuGaR \cite{su2023sugar}, 2DGS \cite{huang2024twodgs}, Gaussian Surfels \cite{dai2024gaussiansurfels}, PGSR \cite{chen2024pgsr}, QGS \cite{zhang2025qgs}, and GOF \cite{yu2024gof}. These methods strengthen surface fitting with priors such as depth regularization, normal regularization, and multi-view consistency. Yet these constraints improve surface fitting only in a fragmented manner: they may refine geometry where observations are strong, still failing to propagate structural constraints across ambiguous regions. During training, explicit geometry rarely enters the optimization loop. In most methods, meshes are extracted only after training via post-processing steps such as TSDF fusion \cite{curless1996volumetric,newcombe2011kinectfusion}. As a result, without a global surface prior to optimize against, Gaussians often drift into artifacts, floaters, and fragmented surfaces in weak-texture or occluded regions.

To bring explicit meshes into the optimization loop, MILo \cite{guedon2025milo} first achieves differentiable mesh extraction during 3DGS training and enforces mutual supervision through rendered depth and normal consistency. Nevertheless, MILo still relies on the basic volumetric Gaussian setting. Its mesh extraction depends on costly local space partitioning and implicit function evaluation. In particular, it introduces virtual corners for each Gaussian to form a local subdivision, computes the implicit values of these corners analytically, and then uses DMTet \cite{shen2021dmtet} to extract the surface. This conversion substantially increases both computational and memory cost. More importantly, it makes the geometric link between discrete Gaussians and continuous meshes less direct.

Compared with volumetric Gaussians, planar Gaussian representations such as 2DGS, Gaussian Surfels, and PGSR are closer to real surfaces and thus better suited for geometry fitting \cite{huang2024twodgs,chen2024pgsr,dai2024gaussiansurfels}. Intuitively, a natural approach is to derive meshes from Gaussian surfels in a differentiable manner, so as to impose global geometric constraints. Yet Gaussian surfels are still essentially an unstructured collection of discrete facets. Although each surfel locally approximates a small tangent patch, the representation as a whole still lacks explicit neighborhood connectivity and manifold structure. It remains challenging to establish an explicit geometric correspondence between a continuous manifold and discrete Gaussian surfels without auxiliary neural networks or extra Gaussian attributes. Accordingly, such a mapping is needed to inject global topological constraints into optimization.

To address this problem, we propose TopoSurfel, a framework for high-fidelity surface reconstruction. We first build a differentiable mesh extraction and rendering pipeline. Considering that surfels capture orientation but do not define explicit front and back sides, we perform normal alignment by combining the global consistency of mesh faces with view-dependent correction, and then sample surfels into a weighted oriented point cloud. We then extract a continuous mesh through geometry-driven differentiable Poisson reconstruction (DPSR) \cite{kazhdan2013screened, peng2021shapeaspoints} and differentiable marching cubes (DiffMC) \cite{lorensen1987marchingcubes,wei2025neumanifold}. The resulting mesh serves as a global explicit geometric prior. By enforcing consistency between the rendered depth and normal maps of Gaussian surfels and meshes, bidirectional supervision can be established.

Building on the extracted proxy mesh, we further propose a mesh-guided surfel evolution strategy. In addition to the classical gradient- and opacity-based density control mechanism, we introduce geometric rules based on explicit point-to-face distances. These rules suppress floaters far from the surface and help fill locally under-covered regions. To better handle background reconstruction in large-scale scenes, we also design a spatially aware hybrid re-initialization strategy. Extensive experiments on multiple benchmarks demonstrate the effectiveness of TopoSurfel in improving geometric reconstruction while preserving rendering quality.

Our main contributions are as follows:

\begin{enumerate}
	\item We propose TopoSurfel, a lightweight framework for surface reconstruction. Without extra learnable parameters, it incorporates a differentiable proxy mesh into Gaussian surfel training and provides a structured surface prior.
	\item Leveraging the extracted mesh prior, we develop a mesh-guided surfel evolution strategy, including normal alignment and geometry-aware density control. The proxy mesh therefore serves not only as supervision but also as a direct guide for surfel structure refinement.
	\item We introduce a stable warm-up stage and a scene-aware re-initialization scheme. Experiments across multiple datasets validate their effectiveness in stabilizing reconstruction for both object-centric and large-scale scenes.
\end{enumerate}

\section{Related Work}
\label{sec:related}

\begin{figure*}[t]
	\Description{Overview of the TopoSurfel Pipeline.}
	\centering
	\includegraphics[width=0.9\textwidth]{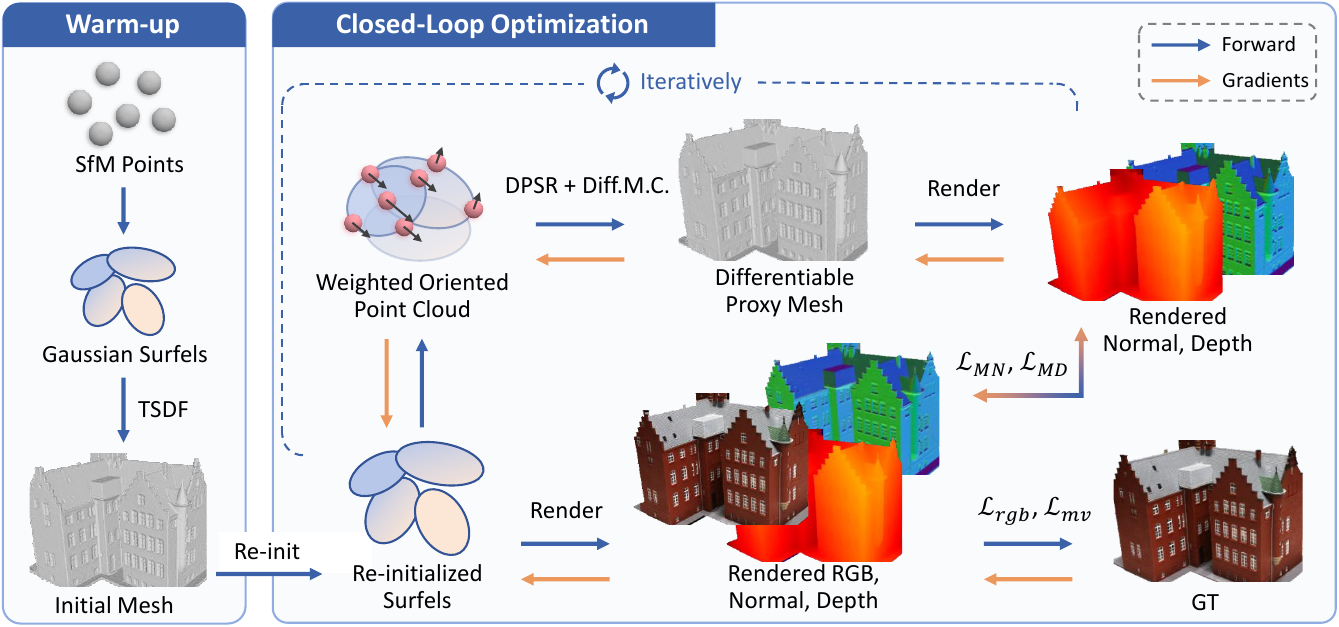}
	\caption{Overview of the TopoSurfel Pipeline. (Left) Warm-up: We initialize Gaussian surfels from SfM points and obtain an initial mesh via TSDF fusion. (Right) Closed-Loop Optimization: In addition to the surfel training pipeline, we extract a weighted oriented point cloud from surfels, and then reconstruct the Differentiable Proxy Mesh from it using DPSR and DiffMC. Both the proxy mesh and the surfels are rendered to compute geometry consistency losses.}
	\label{fig:topo_pipeline}
\end{figure*}

\subsection{Neural Implicit Surface Reconstruction}
The rise of neural radiance fields (NeRF) \cite{mildenhall2020nerf} significantly advanced novel view synthesis based on volume rendering. Follow-up methods such as Deep Blending \cite{hedman2018deepblending} and Instant-NGP \cite{mueller2022instantngp} further explored the limits of computational efficiency and multi-view fusion, while Mip-NeRF 360 \cite{barron2022mipnerf360} extended robust neural rendering to unbounded scenes. However, NeRF represents geometry implicitly as a density field, which often leads to noisy surfaces and incomplete geometric details.

To address this limitation, neural implicit surface reconstruction methods replace density with signed distance fields (SDFs) or occupancy networks. Foundational works such as NeuS \cite{wang2021neus}, VolSDF \cite{yariv2021volsdf}, IDR \cite{yariv2020multiview}, and UNISURF \cite{oechsle2021unisurf} seamlessly combine these implicit representations with volume rendering to enable high-quality surface extraction. Subsequent research further improves reconstruction by either enhancing high-frequency detail recovery \cite{tancik2023neuralangelo,wang2022hfneus} or incorporating external geometric priors and multi-view consistencies \cite{fu2022geoneus,yu2022monosdf}. Despite their strong geometric fidelity, these implicit methods inherently suffer from slow training and rendering speeds, while still requiring complex post-processing for explicit topology extraction.

\subsection{Gaussian-Based Explicit Reconstruction}

In recent years, 3D Gaussian Splatting (3DGS) \cite{kerbl2023gaussian} has become a major research direction because of its explicit point-based representation. While this explicit point-based approach enables high-quality real-time rendering, its unstructured nature presents significant challenges for extracting accurate and continuous surfaces. To enforce better geometric alignment, one primary category of research focuses on refining the rendering primitives. For instance, 2DGS \cite{huang2024twodgs}, Triangle Splatting \cite{held2025trianglesplatting}, and Gaussian Surfels \cite{dai2024gaussiansurfels} replace 3D volumetric Gaussians with 2D disks or triangular facets to better conform to underlying surfaces, while SuGaR \cite{su2023sugar} introduces alignment constraints to facilitate mesh extraction. Building on these foundations, subsequent works like PGSR \cite{chen2024pgsr}, RaDe-GS \cite{zhang2026radegs}, and QGS \cite{zhang2025qgs} further enhance local geometric quality through depth-normal consistency or higher-order curvature modeling. More recently, Geometry-Grounded Gaussian Splatting~\cite{zhang2026geometrygrounded} introduces geometry-grounded depth rendering, while EGG-Fusion~\cite{pan2025eggfusion} explores geometry-aware Gaussian surfel fusion for real-time reconstruction. Unlike online surfel fusion methods, our proxy mesh directly guides normal alignment and density control during optimization.

Alternatively, a distinct category of methodologies explores hybrid representations by bridging explicit Gaussians with implicit fields or external geometric priors. Frameworks such as GOF \cite{yu2024gof}, GSDF \cite{yu2024gsdf}, and 3DGSR \cite{lyu2024_3dgsr} integrate SDF or opacity fields with 3DGS to leverage the continuous nature of implicit representations. However, these approaches typically treat geometry recovery as a decoupled post-processing step or a one-way guidance mechanism. Even recent works on alternative geometry extraction pipelines \cite{wolf2024gs2mesh,stuart2025gs2pc} still keep geometry recovery largely decoupled from Gaussian optimization. As a result, they often lack global topological consistency and struggle to correct geometric errors during optimization, especially in occluded or weak-texture regions.

\subsection{Explicit Geometric Representations and Proxies}
To further improve geometry, several works have explored explicit geometric proxies within the optimization pipeline. For mesh-based representations, nvdiffrec~\cite{munkberg2022nvdiffrec} jointly optimizes geometry, materials, and lighting with a differentiable mesh formulation. MeshSDF~\cite{shang2020meshsdf} and Flexible isosurface extraction~\cite{shen2023flexicubes} also directly formulate differentiable isosurface extraction on explicit geometric proxies. Dmesh~\cite{son2024dmesh} and Dmesh++~\cite{son2025dmeshpp} further improve the fidelity and differentiability of explicit mesh representations, while ExMesh~\cite{fan2026exmesh} directly optimizes explicit meshes with adaptive topology updates. Beyond mesh-based methods, IMLS-Splatting~\cite{yang2025imlssplatting} combines point primitives with implicit moving least squares, while GeoSVR~\cite{li2025geosvr} uses sparse voxels with monocular-depth guidance for geometry reconstruction. These methods show that explicit non-Gaussian representations can provide effective geometric regularization.

\subsection{Gaussian-Mesh Hybrid Reconstruction}
Several recent works have also explored combining Gaussian representations with explicit mesh priors for reconstruction. MILo \cite{guedon2025milo} introduces a mesh supervision pathway during training, while MeshSplatting \cite{held2025meshsplatting} investigates reconstruction with opaque meshes and mesh-based rendering. Although these methods incorporate explicit geometry into the pipeline, the coupling between discrete Gaussians and continuous surfaces remains indirect. In MILo, for example, meshes are obtained through virtual corner subdivision and an intermediate implicit field, followed by DMTet \cite{shen2021dmtet} extraction. This design is end-to-end differentiable, but it still depends heavily on spatial partitioning rather than a direct Gaussian-to-mesh mapping.
\section{Method}
\label{sec:topo_opt}

\subsection{Overview}

As shown in Fig.~\ref{fig:topo_pipeline}, TopoSurfel follows two progressive stages. The warm-up stabilizes Gaussian surfels with photometric and local geometric constraints, and then reinitializes them from a coarse TSDF mesh. In the main surfel-mesh stage, standard surfel training continues while the surfels are differentiably converted into a proxy mesh at every iteration. That mesh provides a structured geometric prior for optimization, while also guiding surfel evolution through normal alignment and geometry-aware density control. This closed loop moves Gaussian primitives from purely photometric fitting toward topology-aware geometric refinement.

\subsection{Spatially-Aware Hybrid Warm-up}
Starting directly from the sparse SfM point cloud is not practical for differentiable closed-loop optimization. At the beginning of training, Gaussians driven only by photometric loss often collapse into a highly scattered ``Gaussian soup'' with little physical connectivity or topological structure. Such a point set also cannot natively support differentiable mesh extraction algorithms such as DPSR \cite{peng2021shapeaspoints}. Forcing mesh extraction at this stage usually introduces severe geometric distortion and can destabilize optimization. We therefore first derive an initial mesh from conventional TSDF fusion \cite{curless1996volumetric, newcombe2011kinectfusion} and then use it to reinitialize the surfels, which provides a stable geometric initialization for the later differentiable mesh extraction stage.

\paragraph{Warm-up.}  We use anisotropic Gaussian primitives to represent the scene, following the standard 3DGS parameterization \cite{kerbl2023gaussian}. Each Gaussian is defined by its center $\mu$, rotation matrix $R$, scale vector $S$, opacity, and spherical harmonic coefficients. During warm-up, in addition to the photometric loss, we continuously compress the shortest axis with scale regularization so that the representation progressively flattens from volumetric Gaussians into surfel-like primitives.

\paragraph{Spatially-Aware Re-initialization.} After warm-up, we perform a spatially aware re-initialization to reinitialize the surfels based on the extracted mesh. We first extract meshes from the rendered depth maps with TSDF fusion \cite{curless1996volumetric,newcombe2011kinectfusion}. Based on this mesh prior, one Gaussian surfel is initialized at the center of each triangle. The face normal and two orthogonal in-plane tangent directions define the local $z$-, $x$-, and $y$-axes of the surfel frame, respectively. The surfel center $\mu$ is set to the triangle centroid, and the rotation matrix $R$ is determined by this local frame. The tangential components of $S$ are estimated from the triangle extent along the corresponding tangents, while the normal component is set to a very small value to preserve a flat shape. The appearance SH coefficients are initialized from the colors of the three triangle vertices. This one-to-one mapping makes the reinitialized surfel set a discrete approximation of the mesh.

\paragraph{Hybrid Re-initialization for Large-scale Scenes.} For large-scale scenes with complex backgrounds, a hybrid initialization strategy is necessary. As TSDF fusion is usually limited by resolution and depth truncation, it often extracts only the central object of the scene. If we simply reset all surfels with this partial mesh, the background surfels would be discarded, leaving no primitives to model the background appearance and causing severe degradation. Therefore, for mesh-covered core regions, we apply the mesh-based reconstruction and normal alignment above; while for uncovered background regions, we simply keep the initially optimized surfels. In practice, we use the nearest-surface distance from each surfel to determine whether it belongs to a mesh-covered region or not. This preserves a clean topological prior for the main observed area while maintaining a complete representation of the background.

\begin{figure}[t]
    \Description{Illustration of mesh-guided normal alignment and geometry-aware density control.}
    \centering
    \includegraphics[width=1.0\linewidth]{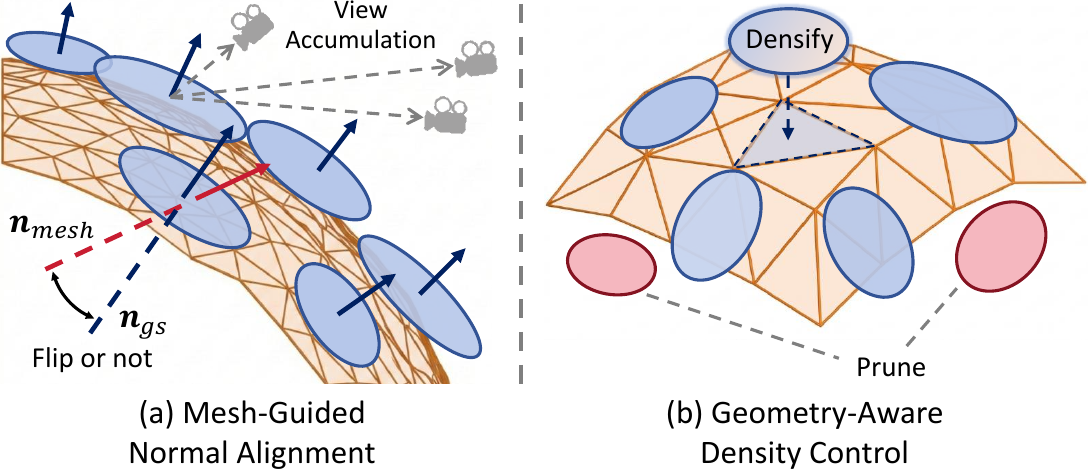}
    \caption{Illustration of Our Mesh-Guided Normal Alignment and Geometry-Aware Density Control}
    \label{fig:method_small}
\end{figure}

\subsection{Differentiable Mesh Extraction}
To introduce global explicit geometry during optimization, we differentiably convert the surfel set into a continuous proxy mesh. This mesh-construction branch introduces no extra learnable parameters and consists of three geometric processing steps.

\paragraph{Weighted Oriented Point Cloud Sampling.} The first step is to build the point cloud required by differentiable Poisson surface reconstruction \cite{kazhdan2013screened,kazhdan2006poisson}. Given the opacity-filtered surfel set $\mathcal{S} = \{s_i \mid \alpha_i > \tau_{\text{opac}}\}$, we regard each surfel as a finite surface patch rather than a point sample. As its center alone cannot faithfully represent the patch extent, we augment each retained surfel with one center point and four symmetric offset samples in its local tangent plane, taken along the two tangential axes at two standard deviations. Specifically, the samples are
\begin{equation}
\mathbf{p}_{i,k} = \boldsymbol{\mu}_i + \mathbf{R}_i(\mathbf{S}_i \odot \mathbf{o}_k), \qquad \mathbf{o}_k \in \{(\pm 2, 0, 0)^T, (0, \pm 2, 0)^T\},
\end{equation}
\noindent where $\odot$ denotes element-wise multiplication. These four samples lie in the local tangent plane orthogonal to the shortest axis. The center point weight is set to $w_{i,0} = \alpha_i$, and the four offset points are assigned $w_{i,k} = 0.5\alpha_i$. All extracted points are then given the oriented surfel normal $\mathbf{n}_i$. Through this dense sampling and weighting strategy, we map the Gaussian parameter set to a physically weighted oriented point cloud $\mathcal{P} = \{(\mathbf{p}_j, \mathbf{n}_j, w_j)\}_{j=1}^{5|\mathcal{S}|}$.

\paragraph{Differentiable Poisson Field Generation.} Once $\mathcal{P}$ is constructed, we use differentiable Poisson surface reconstruction (DPSR) following Shape as Points \cite{peng2021shapeaspoints} to convert it into a continuous 3D scalar field. Concretely, we scatter the point positions $\mathbf{p}_j$ and weighted normals $w_j\mathbf{n}_j$ onto a regular voxel grid through trilinear interpolation, which yields a discrete vector field $\mathbf{V}$. The resulting field is then used to recover the indicator field $\chi$ by solving the Poisson equation in the frequency domain. Let $\widetilde{\mathbf{V}} = \mathrm{FFT}(\mathbf{V})$, the unnormalized frequency-domain solution is
\begin{equation}
\widetilde{\chi}(u) = \widetilde{g}_{\sigma,r}(u) \odot \frac{i\,u \cdot \widetilde{\mathbf{V}}(u)}{-2\pi \|u\|_2^2}, \qquad \chi = \mathrm{IFFT}(\widetilde{\chi}),
\end{equation}
\noindent where $u$ is the spatial frequency and $\widetilde{g}_{\sigma,r}$ is the frequency-domain smoothing kernel. The resulting scalar field is then translated and rescaled to a canonical range. As point rasterization, FFT/IFFT, and the subsequent scalar-field processing are all differentiable, gradients from the mesh branch can be propagated stably back to the surfel parameters through this parameter-free construction.

\paragraph{Explicit Mesh Extraction via DiffMC} Finally, to extract a polygon mesh with explicit topology from the implicit scalar field $\chi$, we apply differentiable marching cubes (DiffMC) following Neumanifold \cite{wei2025neumanifold} and the classical marching cubes formulation \cite{lorensen1987marchingcubes}. Standard marching cubes determines local topology from the signs of the scalar values at voxel corners, which is a discrete and non-differentiable decision process. DiffMC instead models each mesh vertex $\mathbf{v}$ as a linear interpolation of the two voxel-edge endpoints $\mathbf{x}_0$ and $\mathbf{x}_1$ that straddle the iso-surface:
\begin{equation}
\mathbf{v} = \mathbf{x}_0 + \frac{\tau - \chi_0}{\chi_1 - \chi_0}(\mathbf{x}_1 - \mathbf{x}_0)
\end{equation}
\noindent where $\chi_0$ and $\chi_1$ are the scalar values at the two endpoints, and $\tau$ is the iso-surface threshold. This formulation makes the vertex coordinates continuous functions of the underlying scalar field $\chi$. The module finally outputs an explicit proxy mesh $\mathcal{M} = (\mathbf{V}, \mathbf{F})$ consisting of vertices and faces.

\begin{table*}[t]
\centering
\caption{Quantitative geometry comparison on the DTU scenes. Our method achieves the lowest Chamfer Distance without monocular depth priors, while maintaining competitive efficiency. {\color{colorfirst}\raisebox{-0.15mm}{\rule{3mm}{2mm}}} denotes the best, {\color{colorsecond}\raisebox{-0.15mm}{\rule{3mm}{2mm}}} second best, {\color{colorthird}\raisebox{-0.15mm}{\rule{3mm}{2mm}}} third best, respectively. $\dagger$ indicates the use of an additional monocular depth prior.}
\label{tab:dtu_comparison}
\renewcommand{\arraystretch}{1.0}
\setlength{\tabcolsep}{4pt}
\small
\resizebox{\textwidth}{!}{%
\begin{tabular}{c l c c c c c c c c c c c c c c c c c}
\toprule
\multicolumn{2}{l}{CD (mm) $\downarrow$} & 24 & 37 & 40 & 55 & 63 & 65 & 69 & 83 & 97 & 105 & 106 & 110 & 114 & 118 & 122 & Avg. & Time \\
\midrule
\multirow{3}{*}{\rotatebox{90}{\shortstack{NeRF-\\Based}}}
& VolSDF \cite{yariv2021volsdf} & 1.14 & 1.26 & 0.81 & 0.49 & 1.25 & 0.70 & 0.72 & 1.29 & 1.18 & 0.70 & 0.66 & 1.08 & 0.42 & 0.61 & 0.55 & 0.86 & $>$12h \\
& NeuS \cite{wang2021neus}  & 0.83 & 0.98 & 0.56 & 0.37 & 1.13 & 0.59 & 0.60 & 1.45 & 0.95 & 0.78 & 0.52 & 1.43 & 0.36 & 0.45 & 0.45 & 0.77 & $>$12h \\
& Neuralangelo \cite{tancik2023neuralangelo} & 0.45 & 0.74 & 0.33 & 0.34 & 1.05 & \cellcolor{colorthird}0.54 & 0.53 & 1.33 & 1.05 & 0.72 & \cellcolor{colorsecond}0.43 & 0.69 & 0.34 & \cellcolor{colorthird}0.38 & 0.42 & 0.62 & $>$12h \\
\midrule
\multirow{6}{*}{\rotatebox{90}{\shortstack{Gaussian-\\Based}}}
& SuGaR \cite{su2023sugar} & 1.47 & 1.33 & 1.13 & 0.61 & 2.25 & 1.71 & 1.15 & 1.63 & 1.62 & 1.07 & 0.79 & 2.45 & 0.98 & 0.88 & 0.79 & 1.33 & 1h \\
& 2DGS \cite{huang2024twodgs}  & 0.46 & 0.84 & \cellcolor{colorsecond}0.31 & 0.45 & 0.92 & 1.01 & 0.83 & 1.23 & 1.30 & 0.66 & 0.61 & 1.07 & 0.45 & 0.71 & 0.54 & 0.76 & 11m \\
& GOF \cite{yu2024gof}    & 0.50 & 0.82 & 0.37 & 0.37 & 1.12 & 0.74 & 0.73 & 1.18 & 1.29 & 0.68 & 0.77 & 0.90 & 0.42 & 0.66 & 0.49 & 0.74 & 1h \\
& PGSR \cite{chen2024pgsr}  & 0.36 & \cellcolor{colorthird}0.57 & 0.38 & \cellcolor{colorsecond}0.33 & \cellcolor{colorthird}0.78 & 0.58 & \cellcolor{colorthird}0.50 & \cellcolor{colorsecond}1.08 & \cellcolor{colorsecond}0.63 & \cellcolor{colorthird}0.59 & 0.46 & \cellcolor{colorsecond}0.54 & \cellcolor{colorsecond}0.30 & \cellcolor{colorthird}0.38 & \cellcolor{colorsecond}0.34 & \cellcolor{colorthird}0.53 & 30m \\
& QGS  \cite{zhang2025qgs}  & 0.38 & 0.62 & 0.37 & 0.38 & \cellcolor{colorsecond}0.75 & 0.55 & 0.51 & 1.12 & \cellcolor{colorthird}0.68 & 0.61 & 0.46 & \cellcolor{colorthird}0.58 & 0.35 & 0.41 & 0.40 & 0.54 & 48m \\
\midrule
\multirow{2}{*}{\raisebox{0.8ex}{\rotatebox{90}{\footnotesize\shortstack{Other\\Explicit}}}}
& IMLS-Splatting~\cite{yang2025imlssplatting} & \cellcolor{colorfirst}0.32 & 1.32 & 0.67 & 0.62 & 1.16 & 0.80 & 0.78 & 1.45 & 1.06 & 0.89 & 0.67 & 0.97 & 0.63 & 0.63 & 0.64 & 0.84 & 15m \\
& GeoSVR $^\dagger$~\cite{li2025geosvr} & \cellcolor{colorfirst}0.32 & \cellcolor{colorfirst}0.51 & \cellcolor{colorfirst}0.30 & \cellcolor{colorsecond}0.33 & \cellcolor{colorfirst}0.71 & \cellcolor{colorfirst}0.48 & \cellcolor{colorfirst}0.42 & \cellcolor{colorfirst}1.03 & \cellcolor{colorfirst}0.62 & \cellcolor{colorfirst}0.56 & \cellcolor{colorfirst}0.33 & \cellcolor{colorfirst}0.46 & \cellcolor{colorsecond}0.30 & \cellcolor{colorfirst}0.34 & \cellcolor{colorfirst}0.32 & \cellcolor{colorfirst}0.47 & 49m \\
\midrule
\multirow{3}{*}{\rotatebox{90}{\shortstack{Mesh-GS\\Based}}}
& MILo  \cite{guedon2025milo} & 0.43 & 0.74 & 0.34 & 0.37 & 0.80 & 0.74 & 0.70 & 1.21 & 1.22 & 0.66 & 0.62 & 0.80 & 0.37 & 0.76 & 0.48 & 0.68 & 43m \\
& MeshSplatting \cite{held2025meshsplatting} & 0.77 & 0.72 & 0.74 & 0.60 & 0.89 & 1.00 & 0.81 & \cellcolor{colorthird}1.09 & 1.19 & \cellcolor{colorsecond}0.58 & 0.68 & 0.93 & 0.63 & 0.66 & 0.59 & 0.79 & 34m \\
& Ours & \cellcolor{colorthird}0.33 & \cellcolor{colorsecond}0.55 & \cellcolor{colorthird}0.32 & \cellcolor{colorfirst}0.32 & 0.79 & \cellcolor{colorsecond}0.52 & \cellcolor{colorsecond}0.48 & 1.13 & 0.70 & 0.60 & \cellcolor{colorsecond}0.43 & \cellcolor{colorthird}0.58 & \cellcolor{colorfirst}0.29 & \cellcolor{colorfirst}0.34 & \cellcolor{colorthird}0.36 & \cellcolor{colorsecond}0.51 & 37m \\
\bottomrule
\end{tabular}%
}
\end{table*}

\subsection{Mesh-Guided Surfel Evolution}

To guide discrete surfels with the extracted proxy mesh $\mathcal{M}$, we first establish local geometric correspondences between them using a KNN-based nearest-surface search. At each densification step, which also triggers surfel evolution, we compute for every surfel $s_i$ its nearest surface face $f_{\text{near}}^{(i)}$ on $\mathcal{M}$ and record the shortest Euclidean distance $d_i$. This result provides the geometric basis for the subsequent normal alignment and density control modules.

\paragraph{Mesh-Guided Normal Alignment.} To satisfy the oriented point-cloud requirement of differentiable mesh extraction, we design a normal alignment strategy that combines mesh-based cues with local view cues. Unlike GOF~\cite{yu2024gof} and RaDe-GS~\cite{zhang2026radegs}, which estimate 3DGS normals from ray- or rasterization-based formulations, our surfels already provide an intrinsic axis. We therefore only need to determine its orientation. As shown in Fig.~\ref{fig:method_small}, we trust the local mesh prior only when the surfel is sufficiently close to the proxy mesh and its unoriented normal is consistent with the nearest face normal. Let $\hat{\mathbf{n}}_i$ denote the unit normal of the nearest face $f_{\mathrm{near}}^{(i)}$, and let $\mathrm{scale}_i$ denote the in-plane scale of surfel $s_i$. Specifically, the conditions are
\begin{equation}
d_i < \gamma \cdot \mathrm{scale}_i,
\qquad
\left|\mathbf{n}_i \cdot \hat{\mathbf{n}}_i\right| \geq \tau_{\mathrm{cos}},
\end{equation}
where $\gamma$ and $\tau_{\mathrm{cos}}$ control the distance and angular thresholds, respectively. When both conditions are satisfied, we use $\hat{\mathbf{n}}_i$ as the reference direction. Otherwise, we fall back to a view-consistency cue. Whenever surfel $s_i$ is visible at iteration $t$, we update its accumulated viewing-direction vector $\mathbf{v}^{\mathrm{acc}}_i$ using the unit direction from the surfel center $\boldsymbol{\mu}_i$ to the camera center $\mathbf{c}_t$:
\begin{equation}
\mathbf{v}^{\mathrm{acc}}_i \leftarrow \mathbf{v}^{\mathrm{acc}}_i
+ \frac{\mathbf{c}_t - \boldsymbol{\mu}_i}
{\|\mathbf{c}_t - \boldsymbol{\mu}_i\|_2}.
\end{equation}
We define the unified reference direction $\mathbf{v}^{\mathrm{ref}}_i$ as $\hat{\mathbf{n}}_i$ when both conditions above are satisfied, and as $\mathbf{v}^{\mathrm{acc}}_i$ otherwise. We then determine whether to flip the original shortest-axis direction $\mathbf{n}_i$ by
\begin{equation}
\operatorname{sign}_i =
\operatorname{sgn}\!\left(
\mathbf{n}_i \cdot \mathbf{v}^{\mathrm{ref}}_i
\right).
\end{equation}
The oriented normal is given by $\mathbf{n}_i^{*}=\operatorname{sign}_i\mathbf{n}_i$. This operation changes only its orientation and preserves the freedom for subsequent gradient-based refinement.

\paragraph{Geometry-Aware Density Control.} In classical 3DGS, surfel densification and pruning are driven mainly by view-space rendering gradients. This often fails in low-texture, reflective, or occluded regions, leading to holes or floaters. To overcome this limitation, beyond the standard gradient- and opacity-based density control, we add a topology-aware density control strategy based on mesh distance $d_i$ as a strong geometric complement.

In Mesh-Guided Pruning, surfels with low opacities that lie far from the extracted surface, namely $d_i > \beta \cdot D_{\text{scene}} \land \alpha_i < \tau_{\text{prune}}$, are identified as floating artifacts and removed. This explicit physical-distance filter is more decisive than thresholding opacity alone.

In Mesh-Guided Densification, if a face in $\mathcal{M}$ is not the nearest neighbor of any surfel, it reveals a local coverage gap in the reconstructed surface. We therefore add a new surfel at the centroid of that face. The geometric parameters of the new surfel are inherited from the face, following the same rule used in the re-initialization step above, while its SH coefficients are copied from the nearest existing surfel in space. Since the total number of surfels is usually much larger than the number of extracted faces, each valid face typically becomes the nearest neighbor of at least one surfel in well-reconstructed regions. This strategy can fill surface holes precisely and locally without causing an explosion in surfel count.

\begin{table}[t]
\centering
\caption{Quantitative geometry comparison on the TNT dataset.}
\label{tab:tnt_geo}
\renewcommand{\arraystretch}{1.05}
\setlength{\tabcolsep}{2.5pt}
\small
\resizebox{\columnwidth}{!}{%
\begin{tabular}{l|c c|c c c|c c}
\toprule
F1-Score $\uparrow$ & Geo-NeuS & N-angelo & RaDe-GS & PGSR & QGS & MILo (dense) & Ours \\
\midrule
Barn & 0.33 & \cellcolor{colorfirst}0.70 & 0.43 & \cellcolor{colorsecond}0.66 & 0.55 & 0.64 & \cellcolor{colorthird}0.65 \\
Caterpillar & 0.26 & 0.36 & 0.32 & \cellcolor{colorsecond}0.41 & 0.40 & 0.38 & \cellcolor{colorfirst}0.43 \\
Courthouse & 0.12 & \cellcolor{colorsecond}0.28 & 0.21 & 0.21 & \cellcolor{colorsecond}0.28 & \cellcolor{colorfirst}0.31 & 0.22 \\
Ignatius & 0.72 & \cellcolor{colorfirst}0.89 & 0.69 & \cellcolor{colorthird}0.80 & \cellcolor{colorsecond}0.81 & 0.76 & 0.77 \\
Meetingroom & 0.20 & \cellcolor{colorsecond}0.32 & 0.25 & 0.29 & \cellcolor{colorthird}0.31 & 0.28 & \cellcolor{colorfirst}0.34 \\
Truck & 0.45 & 0.48 & 0.51 & \cellcolor{colorthird}0.60 & \cellcolor{colorfirst}0.64 & 0.59 & \cellcolor{colorfirst}0.64 \\
Mean & 0.35 & \cellcolor{colorsecond}0.50 & 0.40 & \cellcolor{colorsecond}0.50 & \cellcolor{colorsecond}0.50 & 0.49 & \cellcolor{colorfirst}0.52 \\
Time & $>$24h & $>$24h & 39m & 71min & 82min & 182m & 98m \\
\bottomrule
\end{tabular}%
}
\end{table}

\subsection{Optimization Objectives}

\paragraph{Initial Local Optimization.} During warm-up, our main goal is to reconstruct scene appearance and make the surfels gradually adhere to the underlying surface. Besides the standard 3DGS photometric loss $\mathcal{L}_{\mathrm{rgb}}$, which combines the $\mathcal{L}_1$ and D-SSIM terms, we add the scale regularization $\mathcal{L}_s$ and the multi-view consistency loss $\mathcal{L}_{\mathrm{mv}}$ following PGSR \cite{chen2024pgsr}. The scale regularization keeps penalizing the shortest axis and forces each Gaussian to flatten into a surfel. While the multi-view loss is designed to improve cross-view geometric alignment by jointly computing the geometric reprojection error and NCC-based photometric consistency between a reference and a neighboring view:
\begin{align}
\mathcal{L}_{\mathrm{mv}} &= \frac{1}{|\mathcal{V}|} \sum_{\mathbf{p} \in \mathcal{V}} \Big(
\|\mathbf{p} - \mathbf{H}_{n \to r} \mathbf{H}_{r \to n} \mathbf{p}\|_2 \notag \\
&\quad + \big(1 - \operatorname{NCC}(\mathbf{I}_r(\mathbf{p}), \mathbf{I}_n(\mathbf{H}_{r \to n}\mathbf{p}))\big)
\Big)
\end{align}
Here $\mathcal{V}$ is the set of valid pixels, $\mathbf{H}_{r \to n}$ is the homography from the reference view to the neighboring view, and $\mathbf{I}(\cdot)$ denotes the rendered color at each pixel.

\begin{table}[t]
\centering
\caption{Quantitative comparison of NVS on the Mip-NeRF 360 dataset.}
\label{tab:mipnerf360_nvs}
\renewcommand{\arraystretch}{1.02}
\setlength{\tabcolsep}{3.5pt}
\small
\resizebox{\columnwidth}{!}{%
\begin{tabular}{l|c c c|c c c}
& \multicolumn{3}{c|}{Indoor scenes} & \multicolumn{3}{c}{Outdoor scenes} \\
& PSNR $\uparrow$ & SSIM $\uparrow$ & LPIPS $\downarrow$ & PSNR $\uparrow$ & SSIM $\uparrow$ & LPIPS $\downarrow$ \\
\hline
NeRF & 26.84 & 0.790 & 0.370 & 21.46 & 0.458 & 0.515 \\
Deep Blending & 26.40 & 0.844 & 0.261 & 21.54 & 0.524 & 0.364 \\
I-NGP & 29.15 & 0.880 & 0.216 & 22.90 & 0.566 & 0.371 \\
Mip-NeRF 360 & \cellcolor{colorfirst}31.72 & 0.917 & 0.180 & \cellcolor{colorsecond}24.47 & 0.691 & 0.283 \\
\hline
3DGS & \cellcolor{colorthird}30.52 & 0.921 & 0.199 & \cellcolor{colorthird}24.45 & \cellcolor{colorthird}0.728 & 0.240 \\
SuGaR & 29.44 & 0.911 & 0.216 & 22.76 & 0.631 & 0.349 \\
2DGS & 30.39 & 0.924 & 0.182 & 24.33 & 0.709 & 0.284 \\
GOF & \cellcolor{colorsecond}30.80 & \cellcolor{colorfirst}0.928 & \cellcolor{colorfirst}0.167 & \cellcolor{colorfirst}24.76 & \cellcolor{colorfirst}0.742 & \cellcolor{colorfirst}0.225 \\
PGSR & 30.35 & \cellcolor{colorsecond}0.924 & \cellcolor{colorthird}0.176 & 24.29 & 0.718 & \cellcolor{colorsecond}0.236 \\
QGS & 30.45 & 0.919 & 0.184 & 24.32 & 0.706 & 0.242 \\
\hline
MILo & 29.13 & 0.916 & 0.186 & 24.23 & \cellcolor{colorsecond}0.740 & 0.264 \\
Ours & 30.47 & \cellcolor{colorthird}0.918 & \cellcolor{colorsecond}0.173 & 24.18 & 0.712 & \cellcolor{colorthird}0.240 \\
\end{tabular}%
}
\end{table}

\paragraph{Surfel-Mesh Co-evolution.} After spatially aware re-initialization, training enters the closed-loop evolution stage. At every optimization step, we extract a continuous proxy mesh $\mathcal{M}$ and render it to obtain the mesh depth map $\mathbf{D}_{\mathrm{mesh}}$ and normal map $\mathbf{N}_{\mathrm{mesh}}$. The depth and normal consistency losses are computed only on valid overlapping pixels. Following MILo \cite{guedon2025milo}, we define depth consistency loss $\mathcal{L}_{\mathrm{MD}}$ and normal consistency loss $\mathcal{L}_{\mathrm{MN}}$ between the surfels and the mesh:
\begin{align}
\mathcal{L}_{\mathrm{MD}} &= \|\log \mathbf{D}_{\mathrm{surfel}} - \log \mathbf{D}_{\mathrm{mesh}}\|_1 \notag \\
\mathcal{L}_{\mathrm{MN}} &= \|\mathbf{N}_{\mathrm{surfel}} - \mathbf{N}_{\mathrm{mesh}}\|_1
\end{align}
The final objective is
\begin{equation}
\mathcal{L} = \mathcal{L}_{\mathrm{rgb}} + \lambda_s \mathcal{L}_s + \lambda_{\mathrm{mv}} \mathcal{L}_{\mathrm{mv}} + \mathbb{I}_{t > T_{\mathrm{wu}}} \big( \lambda_{\mathrm{MD}} \mathcal{L}_{\mathrm{MD}} + \lambda_{\mathrm{MN}} \mathcal{L}_{\mathrm{MN}} \big)
\end{equation}
\noindent where $\lambda_s$, $\lambda_{\mathrm{mv}}$, $\lambda_{\mathrm{MD}}$, and $\lambda_{\mathrm{MN}}$ are the corresponding weights, while $\mathbb{I}_{t > T_{\mathrm{wu}}}$ is an indicator function that activates global mesh supervision after the warm-up stage.

\section{Experiments}
\label{sec:experiments}

\subsection{Implementation Details}
\label{sec:implementation}

We implement TopoSurfel in PyTorch and run all experiments on a single NVIDIA RTX 3090 GPU. The optimization consists of a 10,000 iteration warm-up stage and a 10,000 iteration surfel-mesh co-evolution stage. We use a fixed weight for the scale regularization throughout both the warm-up and the subsequent closed-loop optimization stages. During training, we use nvdiffrast to render the differentiable proxy mesh, which produces depth and normal maps for geometric supervision. After optimization, we use TSDF fusion as the final post-processing step, since higher-resolution DPSR would substantially increase memory consumption and final mesh

\clearpage
\begin{figure*}[t]
\Description{Qualitative geometry comparison on DTU scenes.}
\centering
\includegraphics[width=1\textwidth]{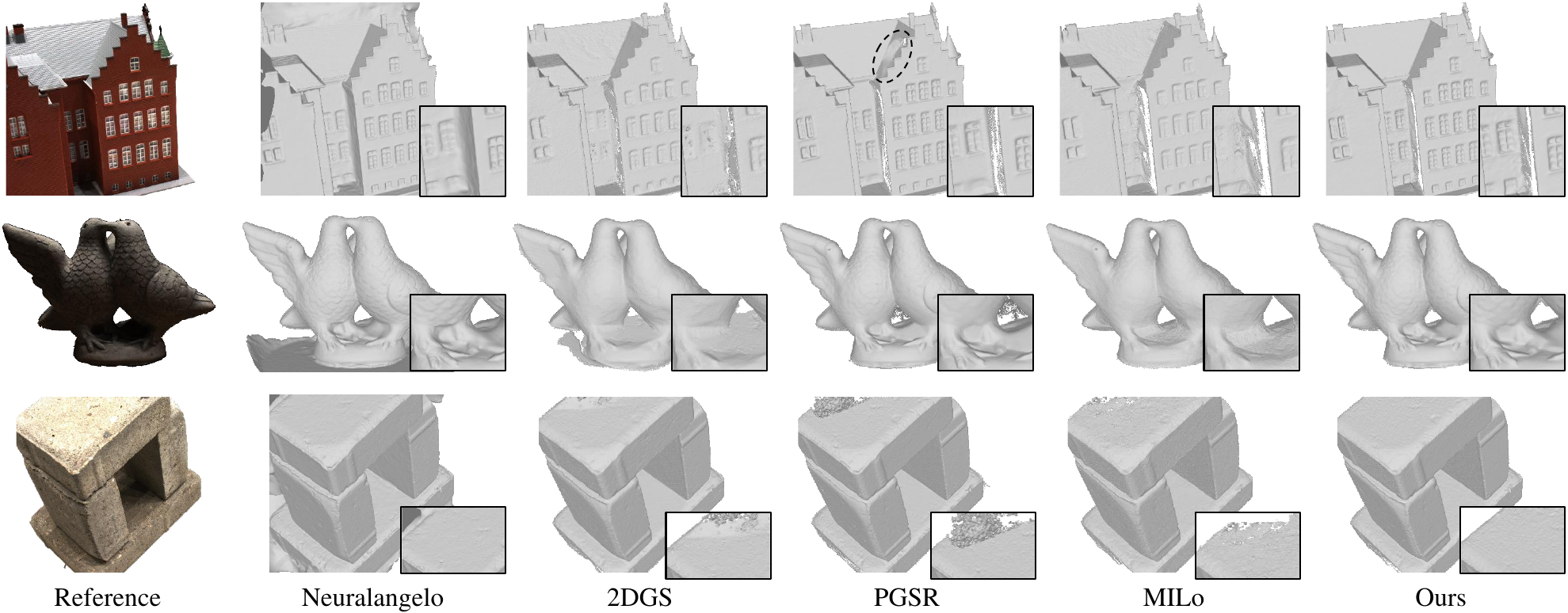}
\vspace{-5mm}
\caption{Qualitative comparison of surface reconstruction on the DTU dataset. In heavily occluded regions and areas with few viewpoints, our method recovers more complete geometry and smoother surfaces with fewer artifacts, whereas other methods often produce over-smoothed or noisy surfaces.}
\label{fig:dtu_geo}
\end{figure*}

\begin{figure*}[t]
\Description{Qualitative geometry comparison on TNT scenes.}
\centering
\includegraphics[width=1\textwidth]{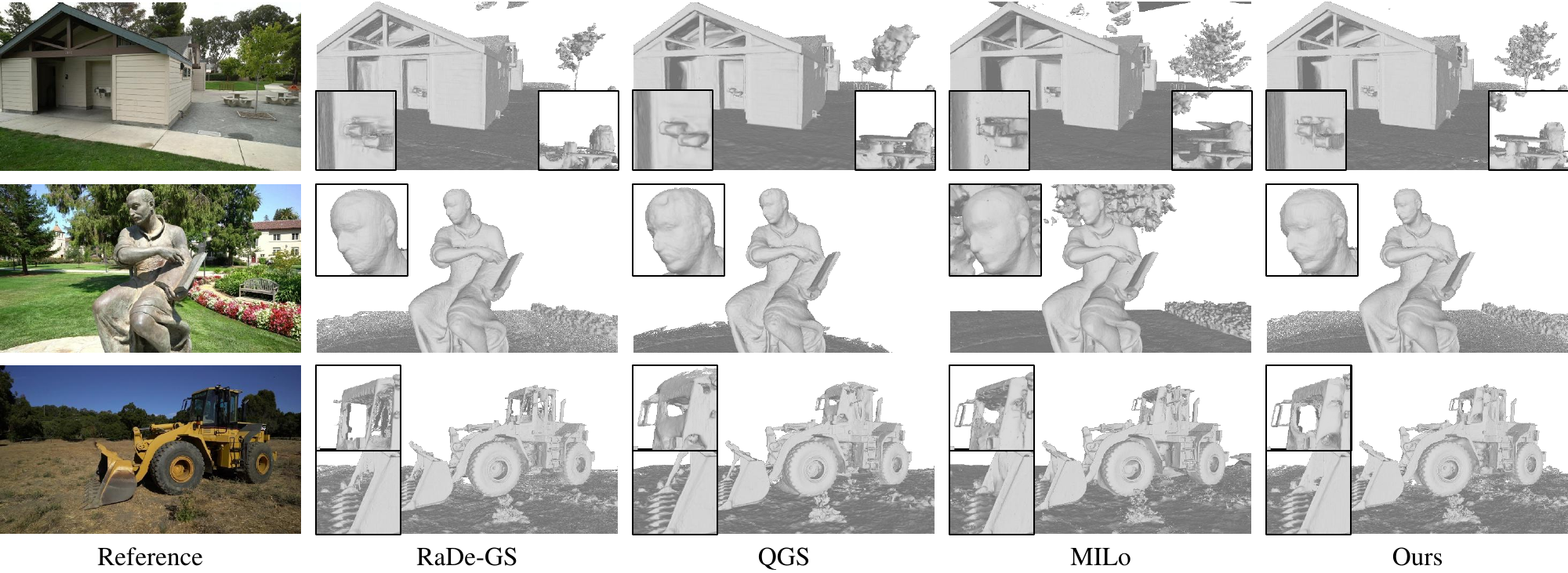}
\vspace{-5mm}
\caption{Qualitative results on the TNT dataset. Our method excels in large-scale real-world scenes, producing cleaner surfaces and more accurate structures. While methods like MILo and QGS struggle with holes or floating noise in complex areas, TopoSurfel maintains high geometric fidelity.}
\label{fig:tnt_geo}
\end{figure*}

\begin{figure*}[t]
\Description{NVS comparison on mipnerf360 scenes.}
\centering
\includegraphics[width=1\textwidth]{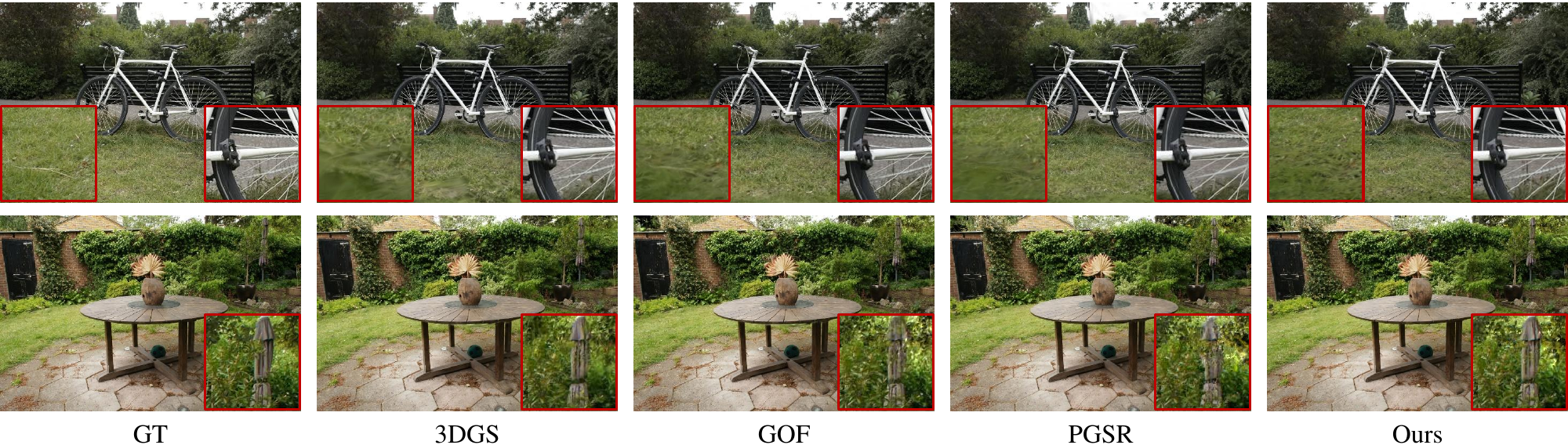}
\vspace{-5mm}
\caption{Novel view synthesis results on the Mip-NeRF 360 dataset. TopoSurfel achieves comparable visual quality to existing methods.}
\label{fig:mipnerf360_nvs}
\vspace{-3mm}
\end{figure*}
\clearpage

\noindent export does not require gradient flow. Additional technical details are provided in the supplementary material.

\begin{table}[t]
\centering
\caption{Quantitative comparison of mesh-based Novel View Synthesis on the NeRF-Synthetic dataset.}
\label{tab:nerf_synthetic_mesh_nvs}
\renewcommand{\arraystretch}{1.05}
\setlength{\tabcolsep}{2pt}
\small
\resizebox{0.92\columnwidth}{!}{%
\begin{tabular}{c|c c c c|c |c c}
\toprule
& 2DGS & GOF & RaDe-GS & PGSR & IMLS-Splat & MILo & Ours \\
\midrule
PSNR $\uparrow$ & 22.10 & \cellcolor{colorsecond}23.97 & 23.58 & 23.14 & 23.72 & \cellcolor{colorthird}23.77 & \cellcolor{colorfirst}25.23 \\
SSIM $\uparrow$ & 0.886 & 0.905 & 0.903 & 0.900 & \cellcolor{colorthird}0.906 & \cellcolor{colorsecond}0.910 & \cellcolor{colorfirst}0.921 \\
LPIPS $\downarrow$ & 0.106 & \cellcolor{colorsecond}0.092 & 0.094 & 0.095 & 0.095 & \cellcolor{colorthird}0.094 & \cellcolor{colorfirst}0.087 \\
$\lvert V \rvert$ & 0.4M & 4.1M & 3.0M & 3.2M & 0.3M & 1.0M & 1.7M \\
\bottomrule
\end{tabular}%
}
\end{table}

\begin{figure}[t]
\Description{Mesh-based novel view synthesis results on NeRF-Synthetic.}
\centering
\includegraphics[width=1\linewidth]{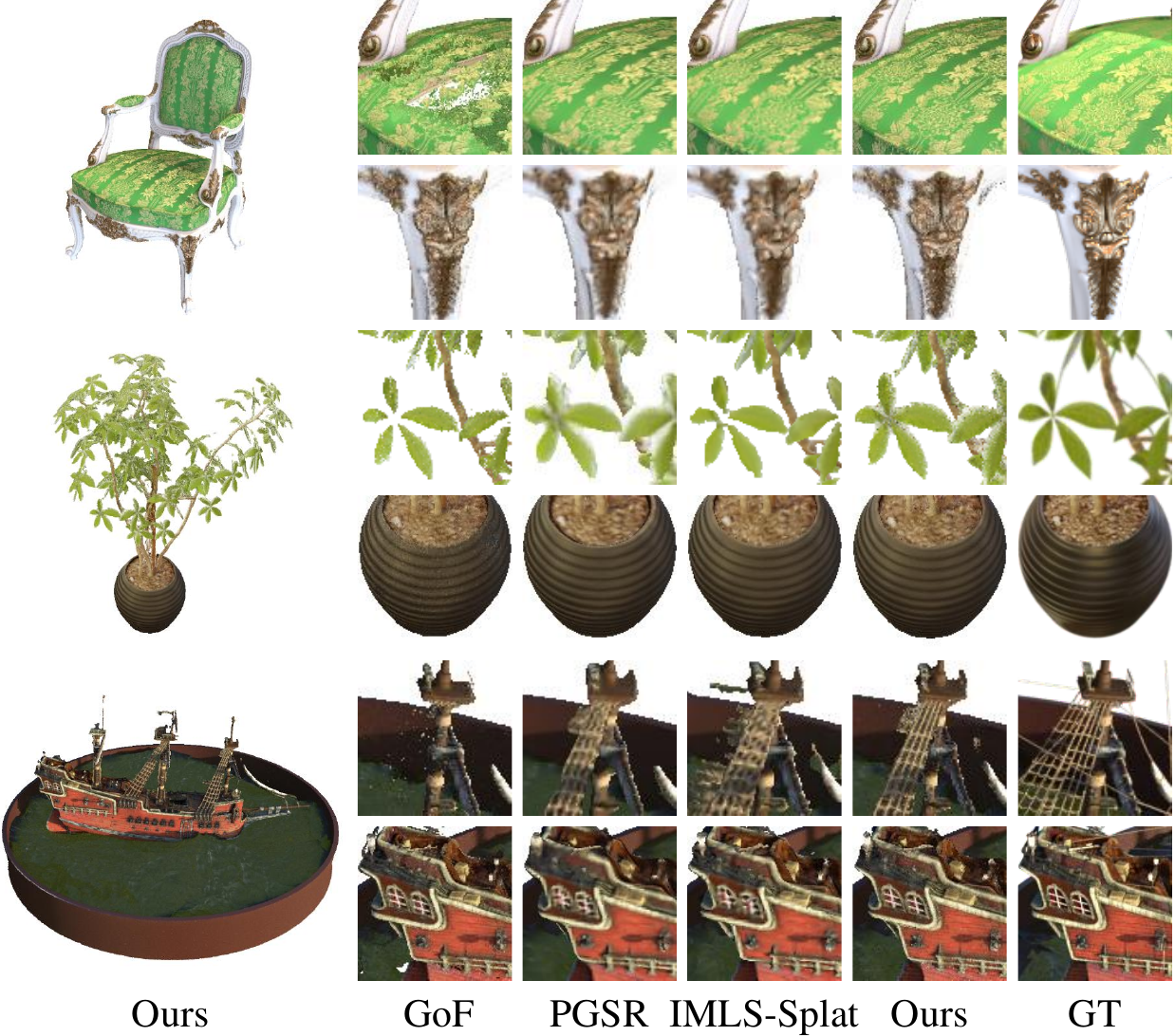}
\caption{Qualitative comparison of Mesh-based NVS on the NeRF-Synthetic dataset. As shown in the zoomed-in regions, our approach preserves intricate details that are often blurred or lost in other methods.}
\label{fig:mesh_based_nvs}
\end{figure}

\subsection{Comparison}
\label{sec:comparison}

\paragraph{Datasets.} We evaluate TopoSurfel on four standard benchmarks. DTU \cite{jensen2014dtu} comprises 15 object-centric scenes captured by structured light and is widely used for quantitative geometry evaluation. Tanks and Temples (TNT) \cite{knapitsch2017tanks} contains 6 complex outdoor scenes with realistic illumination and ground truth point clouds, making it suitable for assessing geometry quality in unbounded environments. Mip-NeRF 360 \cite{barron2022mipnerf360} covers 9 challenging unbounded indoor and outdoor scenes and is mainly used for rendering evaluation. NeRF-Synthetic \cite{mildenhall2020nerf} includes 8 synthetic objects with complex textures and thin structures. We use it for mesh-based NVS evaluation.

\paragraph{Baselines.} We compare TopoSurfel against four categories of methods. (1) NeRF-based methods include NeRF \cite{mildenhall2020nerf}, VolSDF \cite{yariv2021volsdf}, NeuS \cite{wang2021neus}, Neuralangelo \cite{tancik2023neuralangelo}, Geo-NeuS \cite{fu2022geoneus}, Deep Blending \cite{hedman2018deepblending}, Instant-NGP \cite{mueller2022instantngp}, and Mip-NeRF 360 \cite{barron2022mipnerf360}. (2) Gaussian-based methods include 3DGS \cite{kerbl2023gaussian}, 2DGS \cite{huang2024twodgs}, SuGaR \cite{su2023sugar}, GOF \cite{yu2024gof}, RaDe-GS \cite{zhang2026radegs}, PGSR \cite{chen2024pgsr}, and QGS \cite{zhang2025qgs}. (3) Other explicit reconstruction methods include IMLS-Splatting \cite{yang2025imlssplatting} and GeoSVR \cite{li2025geosvr}, where GeoSVR additionally leverages a monocular depth prior. (4) Gaussian-mesh hybrid methods include MILo \cite{guedon2025milo} and MeshSplatting \cite{held2025meshsplatting}. We follow the official settings and evaluation protocols of corresponding methods. For MILo, we use its base configuration on object-level datasets and dense configuration on scene-level datasets. For NeuS, Neuralangelo, and 2DGS, the DTU object masks are additionally provided, as these methods benefit from foreground masking.

\begin{table}[t]
  \centering
  \caption{Comparison of training resources, rendering performance, and output mesh size across methods, where \#GS denotes the number of Gaussians (in millions), GPU Mem denotes peak GPU memory usage (GB), Time denotes training time, FPS denotes frames per second, \#Vertices denotes the number of mesh vertices, and Size denotes mesh file size (MB).}
  \label{tab:computational_efficiency}
  \setlength{\tabcolsep}{2.5pt}
  \small
  \resizebox{\columnwidth}{!}{%
  \begin{tabular}{l | c c c | c | c c}
    \toprule
    & \multicolumn{3}{c|}{Training Resources} & Rendering & \multicolumn{2}{c}{Output Mesh} \\
    & \#GS (M) & GPU Mem & Time & FPS & \#Vertices & Size (MB) \\
    \midrule
    2DGS    & 0.10 & 4.1 GB  & 11m  & 490 & 0.4 M & 15.32 \\
    GOF     & 0.18 & 10.1 GB & 1h  & 41 & 3.8 M & 277.1 \\
    RaDe-GS & 0.13 & 11.3 GB & 14m  & 436 & 2.8 M & 143.7 \\
    PGSR    & 0.16 & 6.4 GB & 30m  & 227 & 3.0 M  & 152.5 \\
    MILo    & 0.20 & 9.4 GB & 43m & 114 & 0.7M  & 41 \\
    Ours    & 0.14 & 8.5 GB & 37m & 284 & 1.6 M  & 83.9 \\
    \bottomrule
  \end{tabular}%
  }
\end{table}

\begin{table}[t]
  \centering
  \caption{Memory stress test under different DPSR grid resolutions on an RTX 3090 GPU with 24 GB memory. We report peak memory, training time, Chamfer distance on DTU ($\downarrow$), and F1-score on TNT ($\uparrow$).}
  \label{tab:memory_stress_test}
  \setlength{\tabcolsep}{4pt}
  \small
  \resizebox{0.9\columnwidth}{!}{%
  \begin{tabular}{c | c c c c}
    \toprule
    Dataset & Grid Resolution & Peak Memory & Time & CD/F1 \\
    \midrule
    \multirow{3}{*}{DTU} & $256^3$ & 4.2 GB & 27m & 0.54 \\
                             & $512^3$ & 8.5 GB & 37m & 0.51 \\
                             & $720^3$ & 17.5 GB & 60m & 0.50 \\
    \midrule    
    \multirow{3}{*}{TNT} & $256^3$ & 9.6 GB & 76m & 0.45 \\
                             & $512^3$ & 14.7 GB & 98m & 0.52 \\
                             & $720^3$ & OOM & -- & -- \\
    \bottomrule
  \end{tabular}
  }
\end{table}

\paragraph{Geometry Evaluation.} On the DTU dataset, we use Chamfer Distance (CD) to measure geometry accuracy. The quantitative results are reported in Table~\ref{tab:dtu_comparison}. TopoSurfel achieves competitive reconstruction quality, outperforms most baselines, and runs at a comparable training cost to current methods. Figure~\ref{fig:dtu_geo} provides the qualitative DTU comparison. In regions with few viewpoints or occlusions, conventional Gaussian regularization methods often produce holes or spurious protrusions. For example, PGSR produces incomplete geometry near the top of the object in the bottom row. In contrast, the global mesh prior helps TopoSurfel recover smoother and more complete surfaces, especially in visually ambiguous regions.

We further evaluate TopoSurfel on the TNT dataset using F1-Score as the geometry metric. As shown in Table~\ref{tab:tnt_geo}, TopoSurfel attains the best geometry accuracy. Although explicit mesh supervision slightly increases training time compared with some Gaussian-based methods, it substantially reduces floaters and surface fragmentation in large-scale scenes. Figure~\ref{fig:tnt_geo} shows the qualitative TNT results. In large-scale scene reconstruction, our method suppresses floaters more effectively and produces cleaner, more coherent meshes.

\paragraph{Rendering Evaluation.} Although TopoSurfel is primarily designed to improve geometry, we also evaluate novel view synthesis (NVS) quality on Mip-NeRF 360 using PSNR, SSIM \cite{wang2004ssim}, and LPIPS \cite{zhang2018lpips}. Table~\ref{tab:mipnerf360_nvs} and Figure~\ref{fig:mipnerf360_nvs} show that our method remains comparable to current SOTA methods in visual fidelity. This indicates that introducing a mesh-based geometric prior can improve surface quality while preserving rendering fidelity.

\begin{table}[t]
\centering
\caption{Ablation study on the TNT dataset.}
\label{tab:tnt_ablation}
\renewcommand{\arraystretch}{1.0}
\setlength{\tabcolsep}{5pt}
\small
\resizebox{0.88\columnwidth}{!}{%
\begin{tabular}{lcc}
\toprule
Setting & F1-Score $\uparrow$ & PSNR $\uparrow$ \\
\midrule
w/o Mesh Loop & 0.42 & 26.75 \\
w/o Warm-up & 0.39 & 26.46 \\
View-Statistics Normal Flips & 0.46 & 26.91 \\
Random Normal Flips & 0.37 & 26.20 \\
w/o Hybrid Initialization & 0.21 & 20.59 \\
w/o Geo-Aware Density Control & 0.48 & 27.03 \\
\midrule
Ours (Full) & 0.52 & 26.83 \\
\bottomrule
\end{tabular}%
}
\end{table}

\begin{figure}[t]
\Description{Ablation of the warm-up stage and mesh-in-the-loop optimization.}
\centering
\includegraphics[width=1\linewidth]{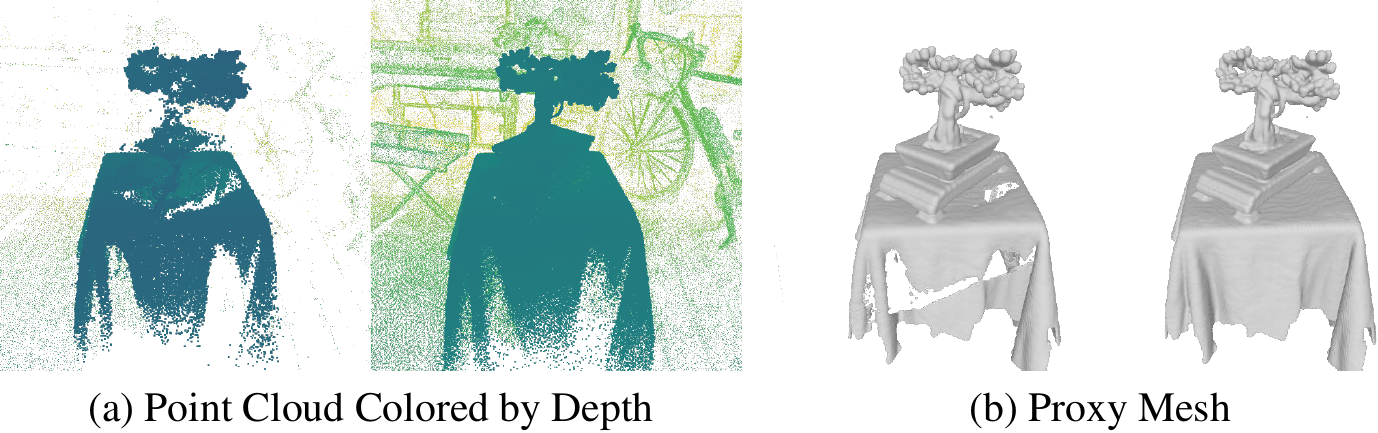}
\caption{Ablation study on the warm-up stage. (a) Gaussian point clouds colored by depth and (b) extracted proxy meshes. In each subfigure, the left side shows the result without warm-up, while the right side shows the result with our warm-up initialization.}
\label{fig:ablation_warmup}
\end{figure}

\paragraph{Mesh-Based NVS} To further assess the usability and completeness of the extracted meshes, we perform mesh-based NVS on NeRF-Synthetic. To eliminate the influence of mesh vertex count across methods on direct rendering quality, the meshes extracted by each method are first UV unwrapped. Then, we optimize only the UV texture maps for 3,000 iterations under photometric supervision from the ground-truth images. At each iteration, the fixed mesh is rasterized with nvdiffrast \cite{laine2020modular}, and colors are sampled from the UV texture maps via the interpolated UV coordinates. Finally, we render novel views using the fixed mesh and optimized UV textures, providing an additional evaluation of mesh usability under a standard texture-based rendering pipeline.

Table~\ref{tab:nerf_synthetic_mesh_nvs} shows that TopoSurfel achieves the best rendering quality. As Figure~\ref{fig:mesh_based_nvs} further shows, for challenging thin structures, complex occlusions, and sparse-view regions, other Gaussian-based methods often produce broken or missing surfaces. This may be attributed to the lack of a strong structural prior, which in turn affects texture mapping. In contrast, our closed-loop optimization can robustly recover fine and complete topological structures.

\begin{figure}[t]
\Description{Ablation of mesh-guided normal alignment.}
\centering
\includegraphics[width=1\linewidth]{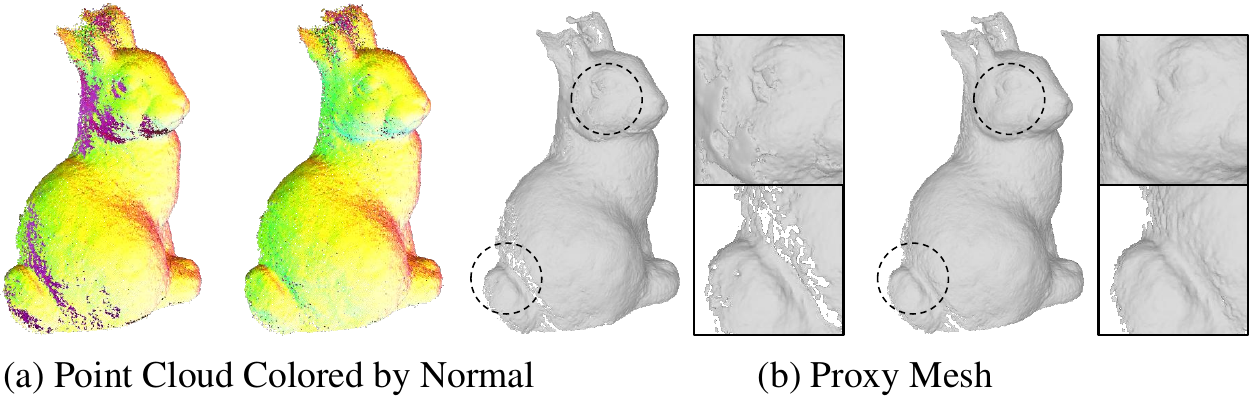}
\caption{Ablation study on mesh-guided normal alignment. (a) Gaussian point clouds colored by surfel normals and (b) extracted proxy meshes. For both (a) and (b), we compare the baseline using only view statistics (left) against our full strategy with mesh-guided normal alignment (right).}
\label{fig:ablation_normal_alignment}
\end{figure}

\begin{figure}[t]
\Description{Ablation of hybrid re-initialization for large-scale scenes.}
\centering
\includegraphics[width=1\linewidth]{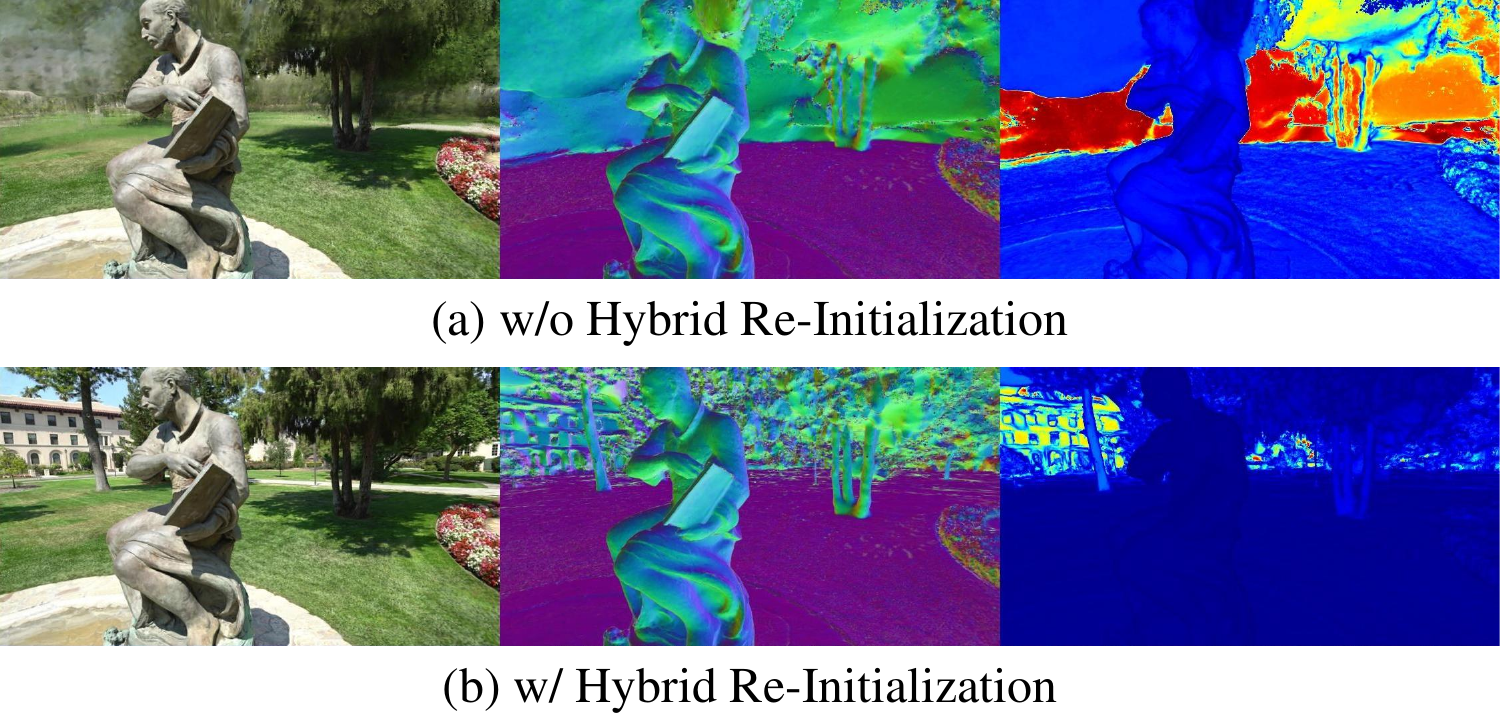}
\caption{Ablation study on hybrid re-initialization for large-scale scenes. (a) Without hybrid re-initialization. (b) With hybrid re-initialization. From left to right, we show the rendered RGB images, normal maps, and depth maps.}
\label{fig:ablation_hybrid_re_init}
\end{figure}

\subsection{Computational Efficiency and Scalability}

\paragraph{Computational Efficiency} TopoSurfel achieves a favorable balance between geometric reconstruction quality and computational cost. Table~\ref{tab:computational_efficiency} reports each method's training resource usage (number of Gaussians, GPU memory, and training time), rendering FPS, and output mesh scale on the DTU dataset, enabling a direct comparison of efficiency and output size.

\paragraph{Memory Stress Test} We further evaluate memory scalability by varying the DPSR grid resolution while keeping the other settings unchanged. As shown in Table~\ref{tab:memory_stress_test}, increasing the resolution from $256^3$ to $512^3$ on DTU yields a clear improvement in Chamfer distance, from 0.54 to 0.51. Further increasing it to $720^3$ brings only a marginal gain to 0.50, while substantially increasing the peak memory and the training time. Scene-level reconstruction on TNT benefits more from the higher DPSR resolution, with the F1-score improving from 0.45 at $256^3$ to 0.52 at $512^3$. However, at $720^3$, three TNT scenes run out of memory on a 24-GB GPU, revealing a current scalability bottleneck. Overall, $512^3$ provides a practical trade-off between reconstruction quality, runtime, and memory consumption.

This limitation stems from the dense volumetric grid used by DPSR, whose memory cost grows cubically with grid resolution. Preserving comparable world-space geometric detail over larger scene extents requires finer grids, making high-resolution proxy extraction increasingly expensive. We therefore use the differentiable mesh as a $512^3$ structural proxy during optimization, while exporting the final mesh through non-differentiable TSDF fusion.

\begin{figure}[t]
\Description{Ablation of geometry-aware density control.}
\centering
\includegraphics[width=1\linewidth]{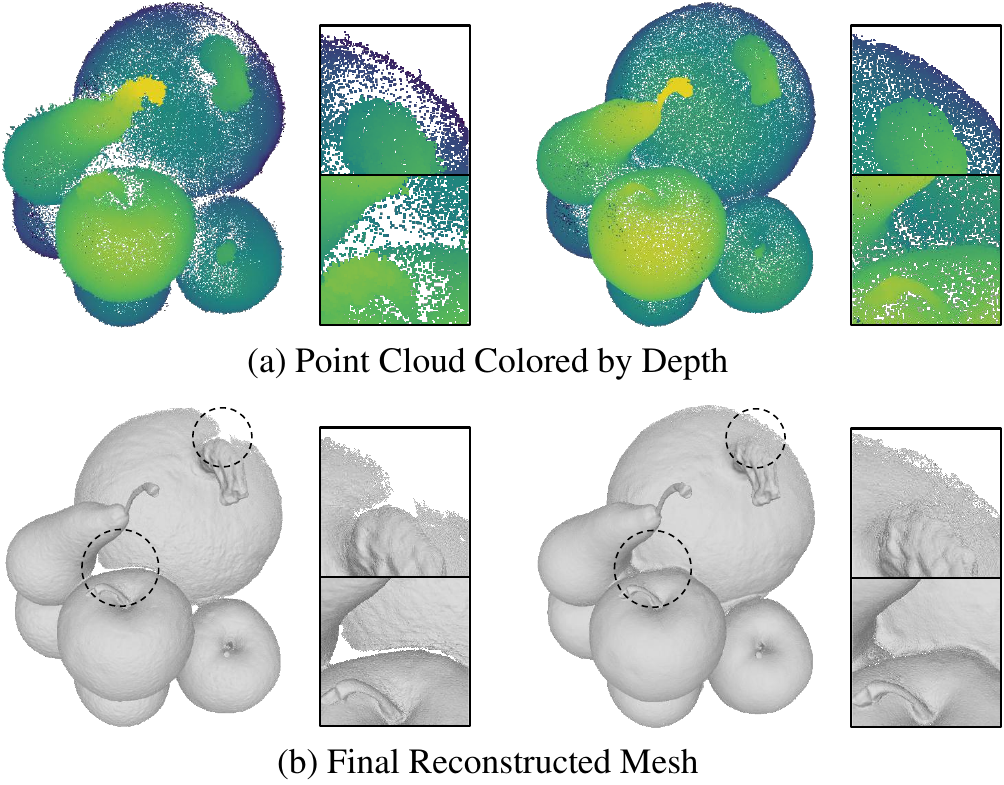}
\caption{Ablation study on geometry-aware density control. (a) Gaussian point clouds colored by depth and (b) final reconstructed meshes. Within each subfigure, the left side uses only gradient-guided density control, whereas the right side incorporates our geometry-aware density control.}
\label{fig:ablation_density_control}
\end{figure}

\subsection{Ablation Study}
\label{sec:ablation}

\paragraph{Effectiveness of Key Pipeline Components.} We first verify the necessity of the core components in our pipeline. Figure~\ref{fig:ablation_warmup} shows that extracting the mesh directly from the coarse SfM point cloud leaves the surfels scattered and produces severe topological distortion. Table~\ref{tab:tnt_ablation} further shows substantial accuracy drops when either the warm-up stage or the mesh loop is removed.

\paragraph{Mesh-Guided Normal Alignment.} For surfel normal estimation, an intuitive approach is to flip normals based on view-direction statistics. As illustrated in Figure~\ref{fig:ablation_normal_alignment}, this view-based strategy can produce incorrect normal directions near boundaries or under sparse viewpoints. As a result, the extracted proxy mesh may become locally broken, which affects both geometry and rendering performance. Moreover, replacing this rule with random flips further degrades performance, as shown in Table \ref{tab:tnt_ablation}. In contrast, our full strategy produces accurate, coherent normals, thereby preserving mesh integrity.

\paragraph{Hybrid Initialization in Large-scale Scenes.} We then evaluate the necessity of the hybrid initialization strategy on large-scale scenes with unbounded backgrounds. After the warm-up stage, reinitializing surfels only from the local TSDF mesh degrades the surfel distribution. As in Figure~\ref{fig:ablation_hybrid_re_init}, this leads to a sharp drop in background depth and rendering quality. Meanwhile, our hybrid initialization combines the foreground mesh with pretrained background surfels to preserve reconstruction stability in large unbounded scenes. The corresponding metrics are summarized in Table~\ref{tab:tnt_ablation}.

\paragraph{Geometry-Aware Density Control.} Finally, we compare the density control that relies only on gradient guidance with our full strategy. As shown in Figure~\ref{fig:ablation_density_control}, classical 3DGS density control is driven solely by view-space gradients, which often creates floaters outside the object and leaves holes in occluded or sparse-view regions. Such an uneven distribution makes the extracted mesh rough and fragmented near the boundaries. Table~\ref{tab:tnt_ablation} reports the corresponding F1-Score and PSNR changes. By contrast, our method produces a smoother and more coherent surface with cleaner edges.

\section{Conclusion}
\label{sec:conclusion}

In this paper, we presented TopoSurfel, a novel framework that bridges discrete Gaussian surfels and continuous meshes. By dynamically extracting a parameter-free proxy mesh and using it to guide surfel evolution, our method achieves accurate and coherent surface reconstruction while maintaining efficient rendering. Despite these advancements, our method has limitations. First, the resolution of the differentiable iso-surfacing is currently bounded by GPU memory capacity, which can limit the extraction of high-frequency micro-geometry in unbounded large-scale scenes compared to offline post-processing. Second, similar to other 3DGS-based approaches, reconstructing highly specular or transparent surfaces remains challenging, as Gaussians tend to model view-dependent radiance rather than underlying actual geometry in these regions. Future work could explore adaptive spatial representations to mitigate memory overhead, and incorporating appearance models to improve the reconstruction of highly reflective surfaces.

\begin{acks}
This work was supported by the Open Fund of National Key Laboratory of Deep Space Exploration (Grant NKDSEL2025008) and the National Natural Science Foundation of China (Grant No. 62306294).
\end{acks}

\bibliographystyle{ACM-Reference-Format}
\bibliography{acmart}


\appendix
\clearpage
\section*{Supplemental Material}
\addcontentsline{toc}{section}{Supplemental Material}

\renewcommand\thesection{\Alph{section}}

\paragraph{Supplementary Material Overview.}
This supplementary document provides implementation details, additional ablation studies, and additional experimental results. The organization is as follows:
\begin{itemize}
    \setlength\itemsep{0.2em}
    \setlength\parsep{0pt}
    \setlength\topsep{0.2em}
    \item \textbf{\hyperref[sec:implementation_details]{Sec.~A: Implementation Details.}}
    Sec.~A.1 provides the pseudocode of mesh-guided normal orientation. Sec.~A.2 describes the hybrid re-initialization scheme for large-scale scenes. Sec.~A.3 details the training and mesh extraction pipeline. Sec.~A.4 summarizes the main hyperparameters.
    \item \textbf{\hyperref[sec:additional_ablations]{Sec.~B: Additional Ablations.}}
    \hyperref[sec:final_mesh_extraction]{Sec.~B.1} studies final mesh extraction with DPSR. \hyperref[sec:diffmc_flexicubes]{Sec.~B.2} compares DiffMC with FlexiCubes as the differentiable mesh extraction operator. \hyperref[sec:number_of_surfel_sampling_points]{Sec.~B.3} evaluates surfel sampling and weighting strategies. \hyperref[sec:normal_estimation_without_gaussian_flattening]{Sec.~B.4} evaluates normal estimation without Gaussian flattening. \hyperref[sec:stagewise_ablation]{Sec.~B.5} analyzes the contribution of each training stage. \hyperref[sec:hyperparameters_ablation]{Sec.~B.6} reports the sensitivity of key hyperparameters.
    \item \textbf{\hyperref[sec:additional_results]{Sec.~C: Additional Results.}}
    \hyperref[sec:dtu_results]{Sec.~C.1} visualizes our reconstructed meshes on DTU. \hyperref[sec:mip360_results]{Sec.~C.2} provides detailed results on Mip-NeRF 360. \hyperref[sec:nerf_synthetic_results]{Sec.~C.3} reports detailed results on NeRF-Synthetic and describes the post-training UV map optimization. \hyperref[sec:proxy_mesh_evolution]{Sec.~C.4} visualizes the evolution of the differentiable proxy mesh during closed-loop optimization. \hyperref[sec:failure_case]{Sec.~C.5} discusses a representative failure case.
\end{itemize}

\section{Implementation Details}\label{sec:implementation_details}
\subsection{Normal Orientation Pseudocode}\label{sec:normal_orientation_pseudocode}
We align surfel normals with a two-stage flip strategy. First, we accumulate view directions from the visible cameras to obtain a provisional orientation. Then, for surfels sufficiently close to the proxy mesh, we trust the local mesh prior and align the normal with the nearest face normal. For surfels farther away from the mesh, we keep the view-consistent orientation. This keeps the orientation rule simple while preserving each surfel's shortest-axis geometry.

\begin{algorithm}[t]
\caption{Mesh-Guided Normal Alignment for TopoSurfel}
\label{alg:normal_alignment}
\begin{algorithmic}[1]
\Require Gaussian surfels $\mathcal{S} = \{\mathbf{s}_i\}_{i=1}^{N}$ with positions $\mathbf{p}_i$ and normals $\mathbf{n}_i$, proxy mesh $\mathcal{M}$, camera views $\mathcal{V}$
\Ensure Consistently oriented normals $\{\mathbf{n}^{*}_i\}_{i=1}^{N}$
\For{each surfel $\mathbf{s}_i \in \mathcal{S}$}
  \State $\mathbf{v}_{\text{ref}} \leftarrow \mathbf{0}$
  \For{each visible view $v \in \mathcal{V}$}
    \State $\mathbf{d}_{i,v} \leftarrow \frac{\mathbf{o}_v - \mathbf{p}_i}{\|\mathbf{o}_v - \mathbf{p}_i\|_2}$ \Comment{$\mathbf{o}_v$ is the camera center}
    \State $\mathbf{v}_{\text{ref}} \leftarrow \mathbf{v}_{\text{ref}} + \mathbf{d}_{i,v}$
  \EndFor
  \State $\mathbf{n}^{\text{init}}_i \leftarrow \operatorname{sign}(\mathbf{n}_i \cdot \mathbf{v}_{\text{ref}})\, \mathbf{n}_i$
  \If{proxy mesh $\mathcal{M}$ is valid and $d_i < \gamma \cdot \text{scale}_i$}
    \State $f_{\text{near}}^{(i)} \leftarrow \text{FindNearestFace}(\mathcal{M}, \mathbf{p}_i)$
    \State $\hat{\mathbf{n}}_i \leftarrow \text{FaceNormal}(f_{\text{near}}^{(i)})$
    \If{$\mathbf{n}^{\text{init}}_i \cdot \hat{\mathbf{n}}_i < 0$}
      \State $\mathbf{n}^{*}_i \leftarrow -\mathbf{n}^{\text{init}}_i$
    \Else
      \State $\mathbf{n}^{*}_i \leftarrow \mathbf{n}^{\text{init}}_i$
    \EndIf
  \Else
    \State $\mathbf{n}^{*}_i \leftarrow \mathbf{n}^{\text{init}}_i$
  \EndIf
\EndFor
\Return $\{\mathbf{n}^{*}_i\}_{i=1}^{N}$
\end{algorithmic}
\end{algorithm}

\begin{table}[h]
\centering
\caption{Hyperparameters used in our implementation.}
\label{tab:supp_hyperparameters}
\small
\renewcommand{\arraystretch}{1.0}
\setlength{\tabcolsep}{10pt}
\resizebox{0.9\columnwidth}{!}{%
\begin{tabular}{l c}
\toprule
Hyperparameter & Value \\
\midrule
\multicolumn{2}{l}{\textit{Optimization hyperparameters}} \\
position\_lr\_init & $1.6 \times 10^{-4}$ \\
position\_lr\_final & $1.6 \times 10^{-6}$ \\
feature\_lr & $2.5 \times 10^{-3}$ \\
opacity\_lr & $5.0 \times 10^{-2}$ \\
scaling\_lr & $5.0 \times 10^{-3}$ \\
rotation\_lr & $1.0 \times 10^{-3}$ \\
$\tau_{\text{prune}}$ & 0.005 \\
\midrule
\multicolumn{2}{l}{\textit{Loss coefficients}} \\
lambda\_dssim & 0.2 \\
multi\_view\_geo\_weight & 0.03 \\
multi\_view\_ncc\_weight & 0.15 \\
mesh\_depth\_weight & 0.05 \\
mesh\_normal\_weight & 0.05 \\
\midrule
\multicolumn{2}{l}{\textit{Geometry thresholds}} \\
Normal-alignment\_cosine & 0.5 \\
Normal\_alignment\_distance $\gamma$ & 3.0 \\
Opacity\_filtering $\tau_{\text{opac}}$ & 0.05 \\
Hybrid\_re-initialization $\beta$ & 0.03 \\
\bottomrule
\end{tabular}%
}
\end{table}

\subsection{Hybrid Re-initialization for Large-Scale Scenes}\label{sec:hybrid_reinitialization}
For large-scale scenes, the TSDF mesh usually covers only the foreground object, because depth truncation and finite resolution limit the reconstructed volume. Reinitializing all surfels from this partial mesh would discard the background and weaken the scene representation. We therefore use a hybrid re-initialization scheme.

In practice, we compute the nearest-surface distance $d_i$ from each surfel to the extracted mesh. A surfel is treated as mesh-covered if $d_i < \beta D_{\text{scene}}$, where $D_{\text{scene}}$ is the scene scale and $\beta$ is a fixed coefficient. The threshold is thus proportional to the scene extent. Surfels in the mesh-covered region are reinitialized from the mesh using the triangle frame described above. Background surfels keep their current parameters and continue training. This keeps the mesh prior on the main object while preserving a complete background.

\subsection{Pipeline Implementation}\label{sec:pipeline_implementation}
\paragraph{Warm-up.} We train the Gaussian surfels for 10,000 iterations in the warm-up stage. Photometric supervision is combined with scale regularization to compress the shortest axis of each Gaussian. This gradually turns the volumetric primitives into surfel-like patches. During warm-up, we reset the Gaussian opacities every 3,000 iterations to prevent early saturation and keep the optimization responsive. We start densification and pruning at iteration 500, and we apply them every 100 iterations until iteration 9,000.

\begin{table}[t]
\centering
\caption{Additional ablation studies on the TNT dataset.}
\label{tab:supp_tnt_ablation}
\renewcommand{\arraystretch}{1.0}
\setlength{\tabcolsep}{5pt}
\small
\resizebox{1.0\columnwidth}{!}{%
\begin{tabular}{lccc}
\toprule
Setting & F1-Score $\uparrow$ & PSNR $\uparrow$ & Time \\
\midrule
\multicolumn{4}{l}{\textit{Mesh extraction}} \\
DPSR final Extraction & 0.43 & 26.83 & 98m \\
FlexiCubes Extraction & 0.39 & 26.67 & 137m \\
\midrule
\multicolumn{4}{l}{\textit{Sampling and weighting}} \\
1-point + opacity & 0.49 & 26.85 & 92m \\
1-point + opacity-area & 0.51 & 26.81 & 96m \\
5-point + KNN density & 0.53 & 26.91 & 142m \\
\midrule
\multicolumn{4}{l}{\textit{Normal estimation}} \\
3DGS + GOF Normal & 0.22 & 24.20 & 129m \\
3DGS + RaDe-GS Normal & 0.39 & 25.59 & 86m \\
\midrule
\multicolumn{4}{l}{\textit{Training stages}} \\
Warm-up only & 0.28 & 25.48 & 5m \\
w/o warm-up & 0.39 & 26.46 & 93m \\
w/o mesh loop & 0.42 & 26.75 & 64m \\
\midrule
\multicolumn{4}{l}{\textit{Hyperparameter sensitivity}} \\
$\gamma$ $0.5\times$ & 0.50 & 26.38 & -- \\
$\gamma$ $2.0\times$ & 0.52 & 26.79 & -- \\
$\beta$ $0.5\times$ & 0.52 & 26.84 & -- \\
$\beta$ $2.0\times$ & 0.52 & 26.77 & -- \\
$\tau_{\text{opac}}$ $0.5\times$ & 0.51 & 26.68 & -- \\
$\tau_{\text{opac}}$ $2.0\times$ & 0.52 & 26.79 & -- \\
\midrule
Ours (Full) & 0.52 & 26.83 & 98m \\
\bottomrule
\end{tabular}%
}
\end{table}

\paragraph{Mesh-Guided Training Schedule.} After the 10,000-iteration warm-up and TSDF-based re-initialization, we enter the 10,000-iteration mesh-guided stage. Within this second stage, proxy mesh supervision is activated from iteration 500. We update the surfel-to-mesh correspondence with a KNN-based nearest-surface search every 100 iterations, using the same interval as densification and pruning. Densification and pruning stop at iteration 7,000 of this stage. In the mesh-guided stage, we no longer reset opacity, since opacity is used to weight the oriented point cloud samples.

\paragraph{Oriented Point Cloud Construction.} At each iteration, we convert the current surfels into a weighted oriented point cloud. We keep surfels with opacity above the preset threshold. For each retained surfel, we sample one center point and four offset points in its local tangent plane. The center point uses the surfel opacity as its weight. The offset points use a smaller weight. All sampled points inherit the surfel normal as their oriented normal. This yields a compact oriented point cloud with five samples per surfel.

\paragraph{DPSR} The oriented points are then fed into DPSR \cite{peng2021shapeaspoints} to obtain a scalar field. Before reconstruction, we normalize the world-space points to the canonical cube $[0,1]^3$ using the current scene center and half-extent. The normalized points and weighted normals are scattered onto a regular voxel grid by trilinear interpolation. We use a voxel resolution of 512 in the training loop. DPSR solves the Poisson equation in the frequency domain and returns a continuous indicator field on the voxel grid. The field is then rescaled to a canonical range for later mesh extraction.

\paragraph{DiffMC} We apply DiffMC \cite{wei2025neumanifold} on the scalar field to extract a differentiable proxy mesh. DiffMC takes the DPSR field as input and uses a fixed iso-surface level of zero. Each mesh vertex is obtained by continuous interpolation between the two voxel-edge endpoints that straddle the iso-surface, making its position differentiable with respect to the scalar field. The resulting proxy mesh is rendered during training for geometric supervision.

\begin{figure}[t]
\Description{Final Mesh Extraction with DPSR and FlexiCubes.}
\centering
\includegraphics[width=1\linewidth]{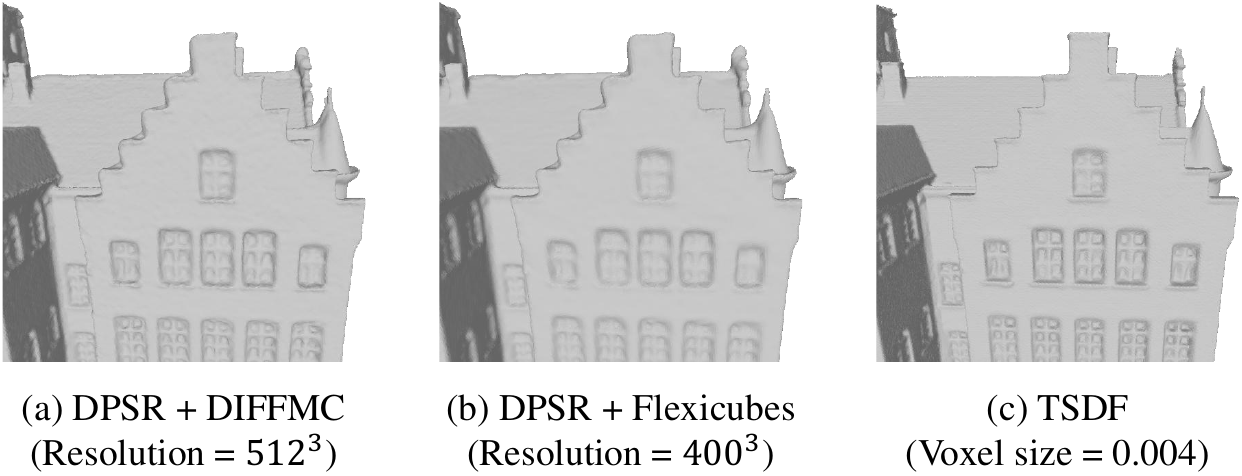}
\caption{Ablation of final mesh extraction with DPSR and FlexiCubes. (a) Mesh extracted with DPSR. (b) Mesh extracted with FlexiCubes. (c) Mesh extracted with TSDF fusion.}
\label{fig:sup_final_mesh_extraction}
\end{figure}

\begin{figure}[t]
\Description{Normal Estimation without Gaussian Flattening.}
\centering
\includegraphics[width=1\linewidth]{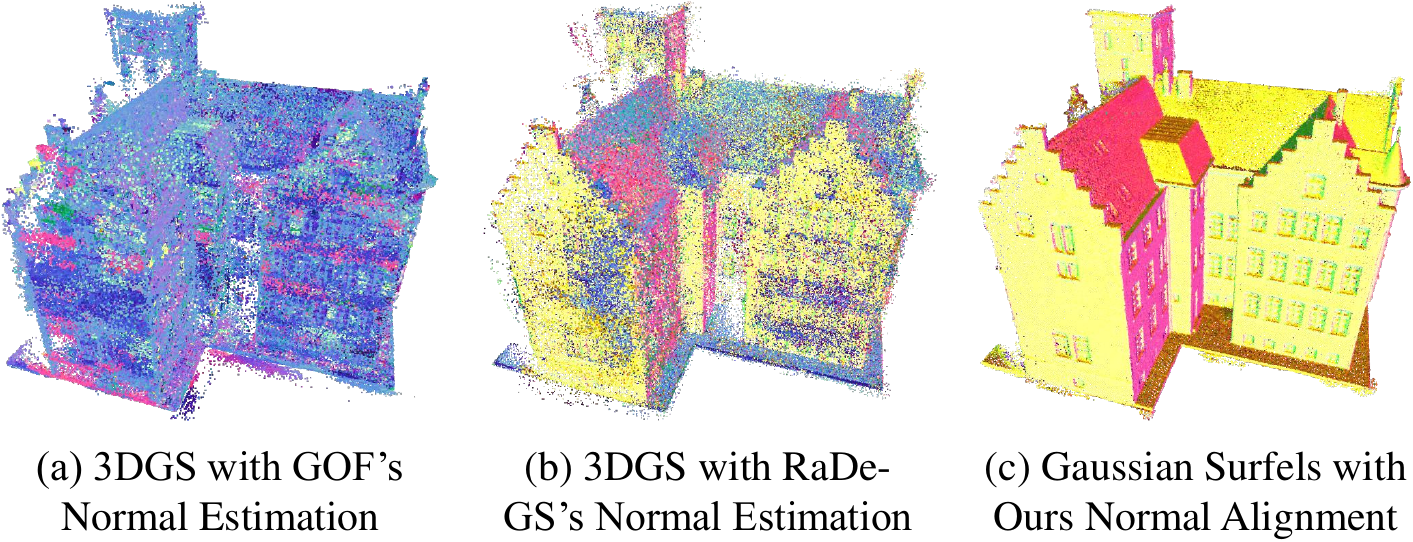}
\caption{Ablation of normal estimation without Gaussian flattening (a) Mesh extracted with GOF's normal estimation. (b) Mesh extracted with RaDe-GS's normal estimation. (c) Mesh extracted with our normal estimation.}
\label{fig:sup_normal_estimation}
\end{figure}

\begin{figure*}[t]
\Description{Visualization of Our Reconstructed Meshes on the DTU Dataset.}
\centering
\includegraphics[width=0.92\textwidth]{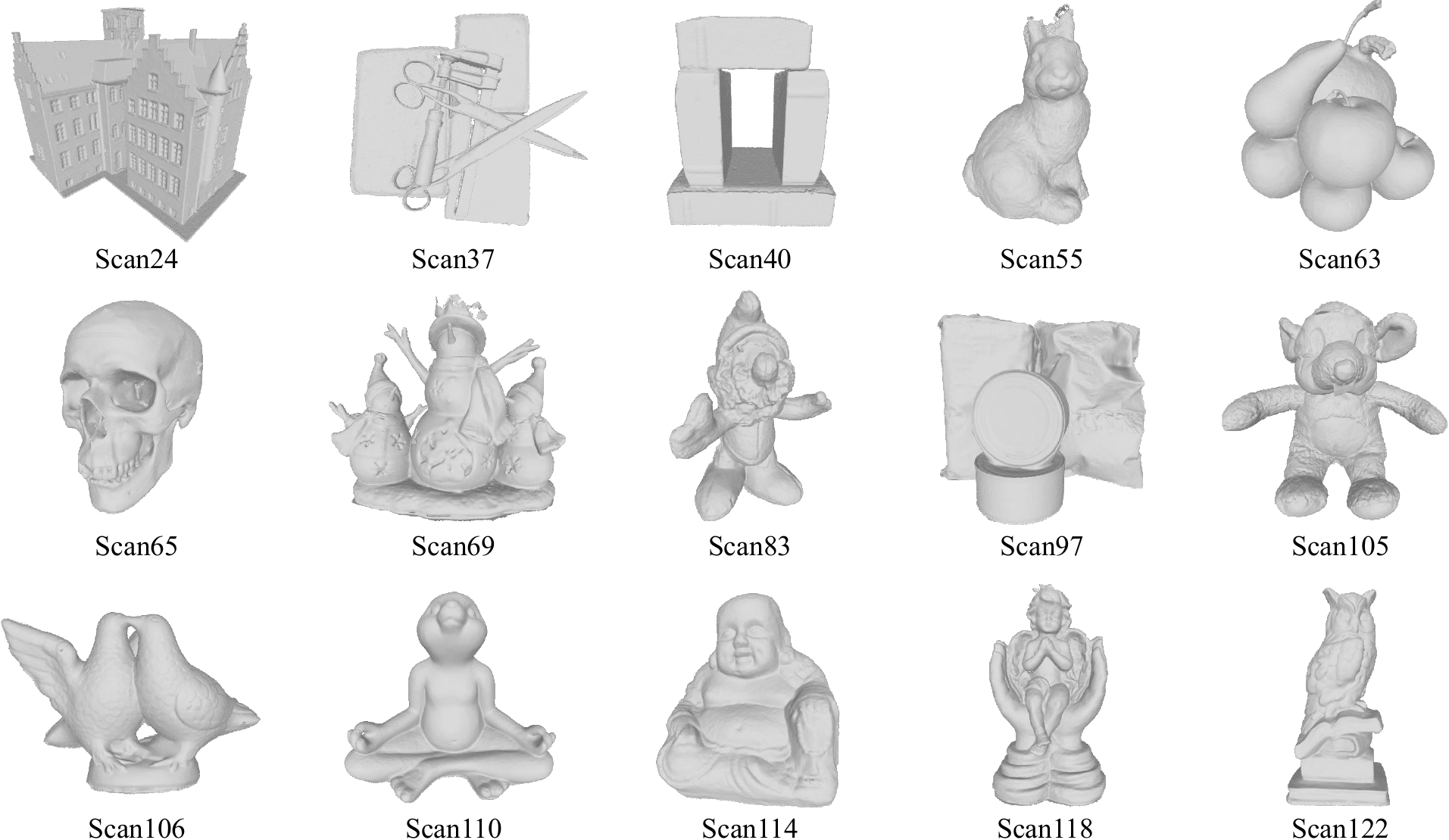}
\caption{Visualization of Our Reconstructed Meshes on the DTU Dataset.}
\label{fig:dtu_ours_mesh}
\end{figure*}

\paragraph{Final Mesh Extraction.} We do not use the proxy mesh as the final output. Instead, we export the reconstructed mesh with TSDF fusion \cite{curless1996volumetric,newcombe2011kinectfusion} after training. This choice is more memory efficient. Increasing the DPSR resolution substantially raises the memory cost. In our experiments, training at a resolution of $720^3$ exceeds the 24-GB GPU memory budget on some scenes. The final mesh export also does not require gradient propagation. Therefore, a non-differentiable TSDF-based extraction is sufficient for the final result.

\subsection{Hyperparameters}\label{sec:hyperparameters_impl}
Table~\ref{tab:supp_hyperparameters} summarizes the main hyperparameters. The same initial learning rates are used for Gaussian parameters in both the warm-up and mesh-guided stages. The notation is consistent across sections: $\tau_{\mathrm{cos}}$ and $\gamma$ control the angular and distance conditions for normal alignment, while $\beta$ controls hybrid re-initialization.

\section{Additional Ablations}\label{sec:additional_ablations}
\subsection{Final Mesh Extraction with DPSR}\label{sec:final_mesh_extraction}
We also evaluate whether the final mesh should be extracted with the same differentiable pipeline used during training, namely DPSR followed by DiffMC. As shown in Fig.~\ref{fig:sup_final_mesh_extraction} and Table~\ref{tab:supp_tnt_ablation}, this alternative is competitive but not ideal as a final output stage. At the default training resolution of 512, the DPSR-based extraction tends to smooth out thin structures and small surface details, which slightly hurts the geometric accuracy on the Tanks and Temples benchmark. Increasing the reconstruction resolution can recover some of these details, but the memory cost grows rapidly because DPSR maintains a dense volumetric field and its intermediate tensors. In our experiments, training at a resolution of $720^3$ exceeds the 24-GB GPU memory budget on some scenes, making this setting impractical for routine use. We therefore use DPSR as a differentiable training proxy and apply TSDF fusion only for final mesh extraction.

\begin{table*}[htbp]
  \centering
  \caption{Quantitative evaluation of novel view synthesis on each scene of the NeRF-Synthetic dataset.}
  \label{tab:quantitative_results_multi_scene}
  \resizebox{\textwidth}{!}{%
  \begin{tabular}{l ccc ccc ccc ccc}
    \toprule
    \multirow{2}{*}{Method} & \multicolumn{3}{c}{Chair} & \multicolumn{3}{c}{Drums} & \multicolumn{3}{c}{Ficus} & \multicolumn{3}{c}{Hotdog} \\
    \cmidrule(lr){2-4} \cmidrule(lr){5-7} \cmidrule(lr){8-10} \cmidrule(lr){11-13}
    & PSNR $\uparrow$ & SSIM $\uparrow$ & LPIPS $\downarrow$ & PSNR $\uparrow$ & SSIM $\uparrow$ & LPIPS $\downarrow$ & PSNR $\uparrow$ & SSIM $\uparrow$ & LPIPS $\downarrow$ & PSNR $\uparrow$ & SSIM $\uparrow$ & LPIPS $\downarrow$ \\
    \midrule
    2DGS & 25.84 & 0.915 & 0.066 & 20.27 & 0.881 & 0.108 & 17.68 & 0.876 & 0.120 & 26.33 & 0.933 & 0.071 \\
    GOF & 26.50 & 0.935 & 0.053 & 21.54 & 0.896 & 0.101 & 23.63 & 0.923 & 0.083 & 27.32 & 0.950 & 0.062 \\
    PGSR & 26.92 & 0.935 & 0.055 & 20.64 & 0.888 & 0.100 & 22.82 & 0.905 & 0.082 & 27.22 & 0.953 & 0.057 \\
    IMLS-Splatting & 27.10 & 0.943 & 0.047 & 21.53 & 0.901 & 0.100 & 23.64 & 0.916 & 0.088 & 28.18 & 0.958 & 0.053 \\
    Ours & 28.01 & 0.943 & 0.056 & 22.50 & 0.914 & 0.085 & 25.19 & 0.940 & 0.068 & 30.52 & 0.967 & 0.051 \\
    \midrule
    \multirow{2}{*}{Method} & \multicolumn{3}{c}{Lego} & \multicolumn{3}{c}{Materials} & \multicolumn{3}{c}{Mic} & \multicolumn{3}{c}{Ship} \\
    \cmidrule(lr){2-4} \cmidrule(lr){5-7} \cmidrule(lr){8-10} \cmidrule(lr){11-13}
    & PSNR $\uparrow$ & SSIM $\uparrow$ & LPIPS $\downarrow$ & PSNR $\uparrow$ & SSIM $\uparrow$ & LPIPS $\downarrow$ & PSNR $\uparrow$ & SSIM $\uparrow$ & LPIPS $\downarrow$ & PSNR $\uparrow$ & SSIM $\uparrow$ & LPIPS $\downarrow$ \\
    \midrule
    2DGS & 23.47 & 0.886 & 0.102 & 18.64 & 0.862 & 0.118 & 22.49 & 0.932 & 0.058 & 22.07 & 0.800 & 0.206 \\
    GOF & 25.37 & 0.919 & 0.075 & 20.76 & 0.869 & 0.123 & 25.03 & 0.952 & 0.052 & 21.60 & 0.794 & 0.188 \\
    PGSR & 24.52 & 0.902 & 0.090 & 17.50 & 0.863 & 0.118 & 23.22 & 0.942 & 0.051 & 22.29 & 0.812 & 0.203 \\
    IMLS-Splatting & 24.38 & 0.903 & 0.091 & 17.58 & 0.852 & 0.144 & 24.11 & 0.948 & 0.055 & 23.22 & 0.810 & 0.179 \\
    Ours & 28.17 & 0.944 & 0.064 & 19.87 & 0.872 & 0.117 & 26.72 & 0.960 & 0.047 & 20.84 & 0.820 & 0.197 \\
    \bottomrule
  \end{tabular}
  }
\end{table*}

\begin{figure*}[t]
\Description{Visualization of Our Reconstructed Meshes and trained UV map on the NeRF-Synthetic Dataset.}
\centering
\includegraphics[width=0.96\textwidth]{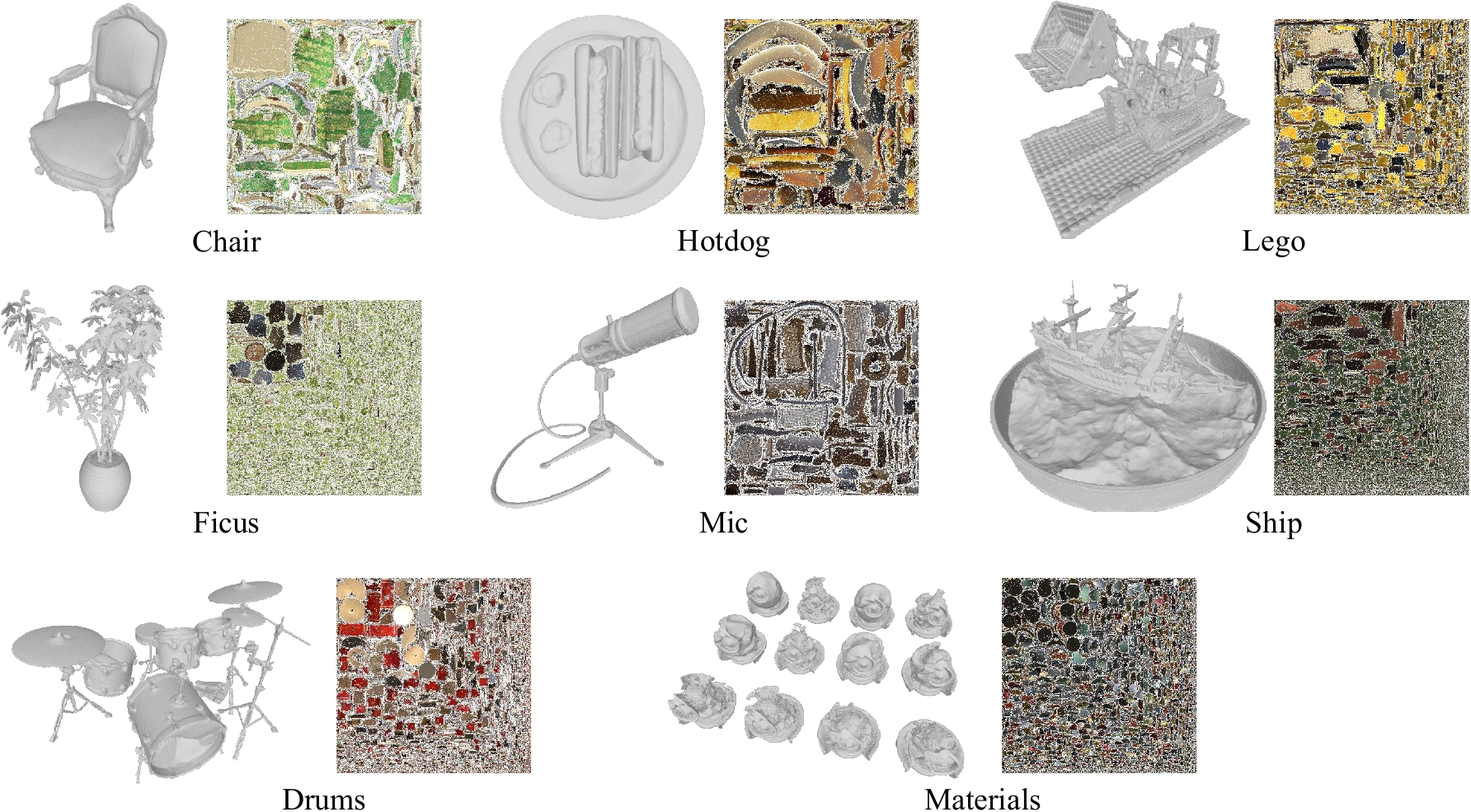}
\caption{Visualization of Our Reconstructed Meshes and trained UV map on the NeRF-Synthetic Dataset.}
\label{fig:nerf_synthetic_ours_mesh}
\end{figure*}

\subsection{DiffMC versus FlexiCubes}\label{sec:diffmc_flexicubes}
We further compare DiffMC with FlexiCubes as the differentiable mesh extraction operator during training. FlexiCubes \cite{shen2023flexicubes} improves the flexibility of surface parameterization by introducing additional weights and vertex offsets, and thus can sometimes produce cleaner topology. However, as reported in NeuManifold \cite{wei2025neumanifold}, this extra expressiveness also increases computation and memory cost. In our setting, running FlexiCubes at a reconstruction resolution of 400 is already close to the hardware limit on a 24~GB GPU, which makes it difficult to scale the extraction resolution further. As a result, the final meshes obtained under this constraint remain less accurate than the meshes produced by our default 512-resolution DiffMC pipeline, as reflected by the results in Fig.~\ref{fig:sup_final_mesh_extraction} and Table~\ref{tab:supp_tnt_ablation}.

\subsection{Sampling and Weighting Strategies}\label{sec:number_of_surfel_sampling_points}
We evaluate the effects of different surfel sampling and weighting strategies on reconstruction quality and efficiency. For single-point sampling, we retain only the surfel center. Opacity weighting uses $w_i=\alpha_i$, while opacity-area weighting uses $w_i=\alpha_i A_i$. We define the surfel area and local KNN density as
\begin{equation}
A_i=S_{i,1}S_{i,2},, \qquad
\rho_i=\left(\frac{1}{K}\sum_{j\in\mathcal{N}_K(i)}
\lVert\boldsymbol{\mu}_i-\boldsymbol{\mu}_j\rVert_2^2+\epsilon\right)^{-1},
\end{equation}
where $S_{i,1}$ and $S_{i,2}$ are the two tangential scales, $\mathcal N_K(i)$ denotes the set of $K$ nearest neighboring surfels, and $\epsilon$ is a small constant for numerical stability. For KNN-density weighting, we scale the default five-point weights inversely with the local density, using $w_{i,0}=\alpha_i/\rho_i$ for the center and $w_{i,k}=0.5\alpha_i/\rho_i$ for the four offset samples. Related density- and area-aware formulations have also been explored for point-cloud surface reconstruction~\cite{barill2018fast,chen2024dipole}. As shown in Table~\ref{tab:supp_tnt_ablation}, Single-point opacity weighting yields an F1-score of 0.49, while incorporating surfel area improves it to 0.51. Five-point KNN-density weighting achieves the highest F1-score of 0.53, but increases the training time from 98 to 142 minutes. We therefore retain five-point opacity weighting in the full model, as it provides a favorable balance between reconstruction quality and training efficiency in practice.

\subsection{Normal Estimation without Gaussian Flattening}\label{sec:normal_estimation_without_gaussian_flattening}
As shown in Fig.~\ref{fig:sup_normal_estimation} and Table~\ref{tab:supp_tnt_ablation}, we study a variant that removes the Gaussian Flattening Loss and estimates normals only from the covariance eigenvectors, following the idea used in GOF \cite{yu2024gof} and RaDe-GS \cite{zhang2026radegs}. We keep the five-point sampling strategy unchanged, but we no longer constrain each Gaussian to collapse toward a local tangent plane. In practice, this makes the normal direction less stable in topologically complex regions. The resulting proxy mesh becomes noisier and less consistently oriented. More importantly, because the mesh-guided loop relies on this proxy mesh to provide geometric supervision, the degraded mesh can propagate incorrect constraints back to the surfels. This not only lowers the final reconstruction accuracy, but also harms rendering quality due to the noisy geometric guidance.

\begin{figure}[t]
\Description{Evolution of the differentiable proxy mesh during closed-loop optimization.}
\centering
\includegraphics[width=1\linewidth]{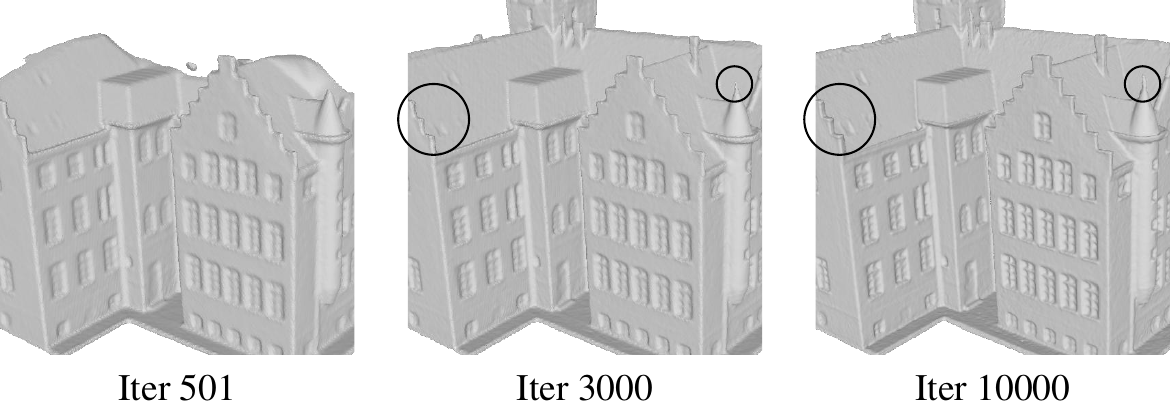}
\vspace{-5mm}
\caption{Evolution of the differentiable proxy mesh during closed-loop optimization. The proxy mesh is used as a training-time geometric prior. Iterations are counted from the beginning of the mesh-guided stage.}
\label{fig:proxy_mesh_evolution}
\end{figure}

\begin{figure}[t]
\Description{Failure Case.}
\centering
\includegraphics[width=0.95\linewidth]{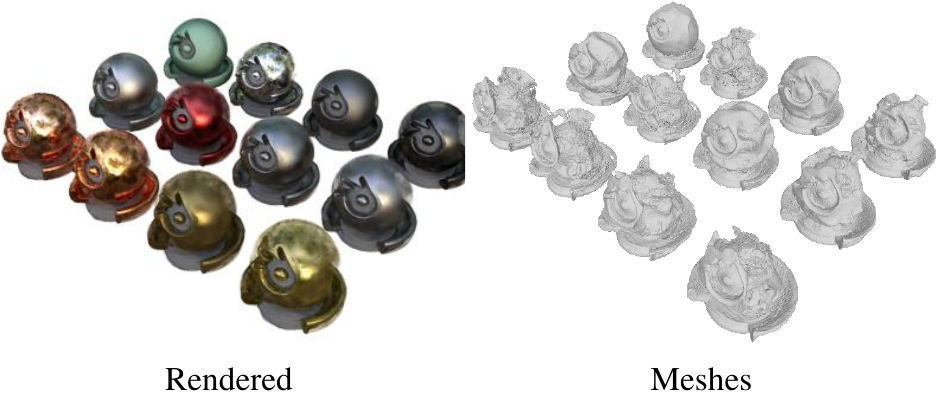}
\vspace{-1mm}
\caption{Failure case on the Materials scene of NeRF-Synthetic.}
\label{fig:failure_case}
\vspace{-5mm}
\end{figure}

\subsection{Stage-wise Ablation}\label{sec:stagewise_ablation}
We evaluate the contributions of the warm-up stage and the surfel-mesh optimization loop on TNT. The initial TSDF mesh extracted after the warm-up stage provides only a coarse reconstruction, reaching an F1-score of 0.28. Directly entering closed-loop optimization without warm-up saves only five minutes, but reduces the F1-score from 0.52 to 0.39, demonstrating the importance of stable initialization. Removing the surfel-mesh loop reduces the training time from 98 to 64 minutes, but also decreases the F1-score to 0.42, while affecting PSNR only slightly. These results show that the warm-up stage and mesh-guided co-evolution play complementary roles: the former provides a reliable initialization, whereas the latter primarily improves geometric reconstruction.

\subsection{Hyperparameters}\label{sec:hyperparameters_ablation}
As shown in Table~\ref{tab:supp_tnt_ablation}, the mesh-guided normal alignment threshold $\gamma$ has the clearest effect among the tested hyperparameters. When $\gamma$ is reduced, fewer surfels are allowed to trust the local mesh prior. More surfels then rely on accumulated view directions for normal estimation, which makes the normals less accurate in difficult regions and leads to a clear drop in reconstruction quality. In contrast, the performance is much less sensitive to the hybrid re-initialization threshold $\beta$ and the opacity filtering threshold $\tau_{\text{opac}}$. Although these two parameters affect which surfels are reinitialized or retained, the subsequent densification and pruning process, together with opacity learning, can largely compensate for moderate changes in their values. This makes the overall reconstruction accuracy stable across a reasonable range of settings.

\section{Additional Results}\label{sec:additional_results}
\subsection{Our Reconstructed Meshes on DTU}\label{sec:dtu_results}
DTU \cite{jensen2014dtu} is a standard multi-view stereo benchmark with controlled capture conditions and accurate ground truth geometry. Figure~\ref{fig:dtu_ours_mesh} shows the complete set of reconstructed meshes on DTU. These results provide an additional qualitative view of the meshes produced by our method on the DTU scenes.

\subsection{Detailed Results on Mip-NeRF 360}\label{sec:mip360_results}
\begin{figure*}[t]
  \Description{Comprehensive visualization of results on multiple Mip-NeRF 360 scenes, showing rendered images, reconstructed meshes, Surfel normals, and mesh normals.}
  \centering
  \includegraphics[width=1\textwidth]{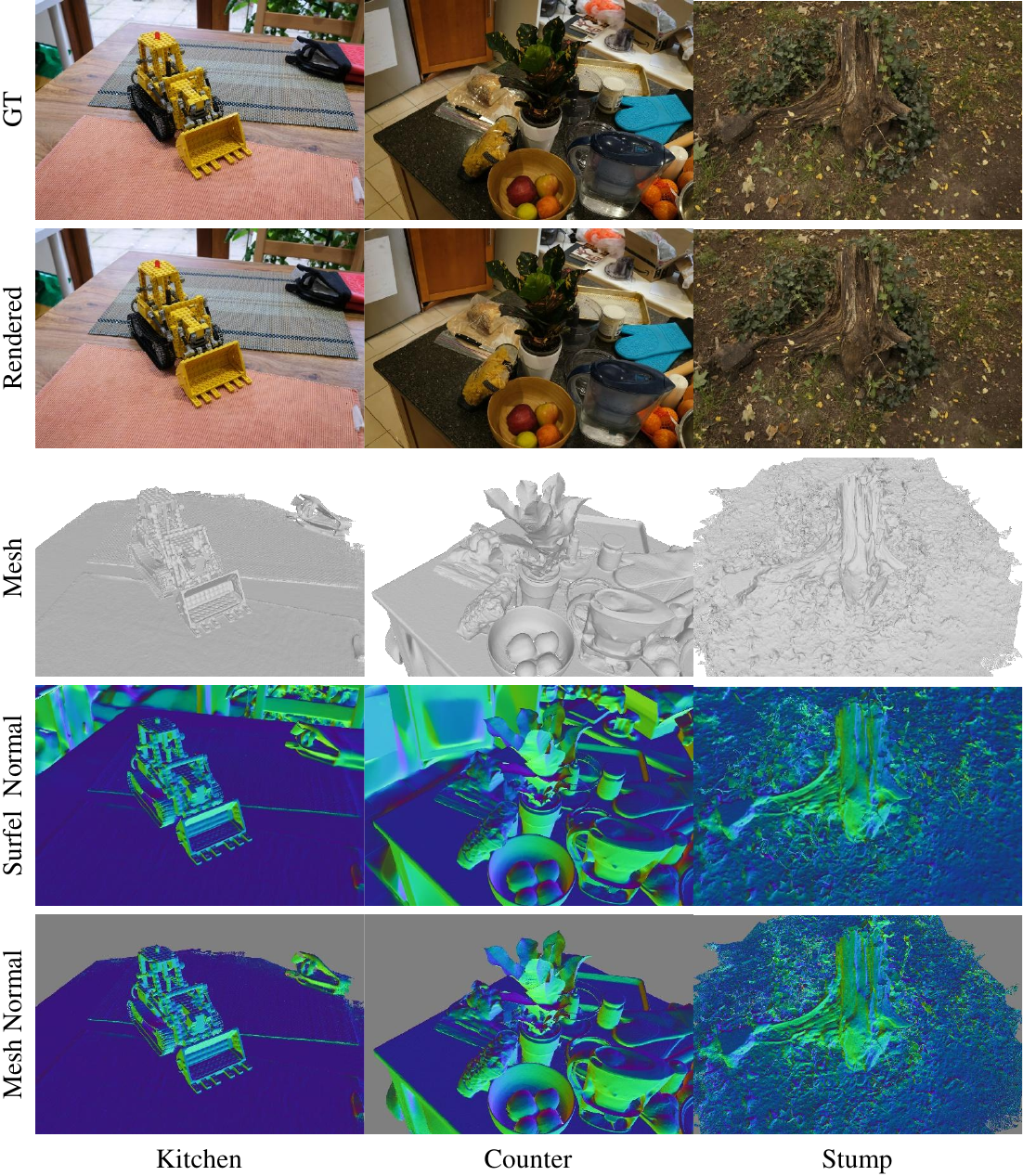}
  \caption{Comprehensive visualization of results on multiple Mip-NeRF 360 scenes, including ground-truth images, our rendered images, reconstructed meshes, Surfel normals, and mesh normals.}
  \label{fig:mip360_comprehensive}
\end{figure*}

For the Mip-NeRF 360 dataset, Fig.~\ref{fig:mip360_comprehensive} presents the reconstructed meshes together with rendered images, Surfel normals, and mesh normals. This dataset contains unbounded outdoor and indoor scenes with more complex backgrounds and illumination changes. The visualization shows that our extracted meshes remain geometrically consistent with the rendered appearance.

\subsection{Detailed Results on NeRF-Synthetic}\label{sec:nerf_synthetic_results}
For the NeRF-Synthetic dataset \cite{mildenhall2020nerf}, we also provide additional details on the post-training UV map optimization. We use xatlas to unwrap the reconstructed mesh and generate a $2048 \times 2048$ texture atlas. The texture map is initialized from the mesh vertex colors. We then fix the geometry and optimize only the texture parameters for 3,000 iterations using photometric loss (within 3 minutes), with a texture learning rate of 0.005. This design reduces the dependence of rendering quality on mesh vertex density and enables the texture map to capture finer appearance details.

Table~\ref{tab:quantitative_results_multi_scene} presents the detailed per-scene mesh-based NVS results on the NeRF-Synthetic dataset. In addition, we show the reconstructed meshes together with post-training UV maps in Fig.~\ref{fig:nerf_synthetic_ours_mesh}.

\subsection{Visualization of Proxy Mesh Evolution}\label{sec:proxy_mesh_evolution}
We further visualize the evolution of the differentiable proxy mesh during closed-loop optimization. As shown in Fig.~\ref{fig:proxy_mesh_evolution}, the proxy mesh is already initialized after the warm-up stage but still contains incomplete structures at the beginning of mesh-guided training. As optimization proceeds, the mesh becomes progressively more coherent: spurious components are suppressed, surface holes are reduced, and fine architectural structures such as window boundaries and facade details become more stable. This behavior indicates that the proxy mesh does not merely provide a static intermediate representation, but continuously co-evolves with the Gaussian surfels and provides increasingly reliable geometry guidance.

\subsection{Failure Case}\label{sec:failure_case}
As shown in Fig.~\ref{fig:failure_case}, we present a rendering result and the reconstructed mesh on the \textit{Materials} scene of NeRF-Synthetic. Similar to many 3DGS-based methods \cite{huang2024twodgs,yu2024gof,chen2024pgsr,zhang2026radegs}, our approach still struggles with highly reflective metallic surfaces. In such regions, strong view-dependent appearance and weak geometric cues can interfere with both surfel optimization and mesh extraction, which may lead to incomplete surface recovery and less stable reconstruction quality.

\end{document}